\documentclass{article}

\usepackage{arxiv}

\usepackage[utf8]{inputenc} % allow utf-8 input
\usepackage[T1]{fontenc}    % use 8-bit T1 fonts
\usepackage{hyperref}       % hyperlinks
\usepackage{url}            % simple URL typesetting
\usepackage{booktabs}       % professional-quality tables
\usepackage{amsmath,amssymb,amsfonts}
\usepackage{nicefrac}     
\usepackage{microtype} 
\usepackage{graphicx}
\usepackage{doi}
\usepackage{authblk}
\usepackage{chngcntr}
\usepackage{makecell}
\usepackage{longtable}
\usepackage{subcaption}
\usepackage{multirow}
\usepackage[style=numeric,sorting=none]{biblatex}

\title{Foundation Model of Electronic Medical Records for Adaptive Risk Estimation}

\author[1,2,3]{Pawel Renc}
\author[1,2]{Michal K. Grzeszczyk}
\author[4]{Nassim Oufattole}
\author[5,2]{Deirdre Goode}
\author[4]{Yugang Jia}
\author[6]{Szymon Bieganski}
\author[2]{Matthew B. A. McDermott}
\author[3]{Jaroslaw Was}
\author[1,2]{Anthony E. Samir}
\author[7,2]{Jonathan W. Cunningham}
\author[7,8,2]{David W. Bates}
\author[1,2,*]{Arkadiusz Sitek}

\affil[1]{Massachusetts General Hospital, Boston, USA}
\affil[2]{Harvard Medical School, Boston, USA}
\affil[3]{AGH University of Krakow, Krakow, Poland}
\affil[4]{Massachusetts Institute of Technology, Cambridge, USA}
\affil[5]{Newton Wellesley Hospital, Newton, USA}
\affil[6]{Medical University of Lodz, Lodz, Poland}
\affil[7]{Brigham and Women’s Hospital, Boston, USA}
\affil[8]{Harvard Chan School of Public Health, Boston, USA}
\affil[*]{\textit{Corresponding author: sarkadiu@gmail.com}}

\date{}

\renewcommand{\headeright}{}
\renewcommand{\undertitle}{}
\renewcommand{\shorttitle}{Foundation Model of EMR for Adaptive Risk Estimation}

\begin{document}
\maketitle

\begin{abstract}
Hospitals struggle to predict critical outcomes. Traditional early warning systems, like NEWS and MEWS, rely on static variables and fixed thresholds, limiting their adaptability, accuracy, and personalization. We previously developed the Enhanced Transformer for Health Outcome Simulation (ETHOS), an AI model that tokenizes patient health timelines (PHTs) from EHRs and uses transformer-based architectures to predict future PHTs. ETHOS is a versatile framework for developing a wide range of applications. In this work, we develop the Adaptive Risk Estimation System (ARES) that leverages ETHOS to compute dynamic, personalized risk probabilities for clinician-defined critical events. ARES also features a personalized explainability module that highlights key clinical factors influencing risk estimates. We evaluated ARES using the MIMIC-IV v2.2 dataset together with its Emergency Department (ED) extension and benchmarked performance against both classical early warning systems and contemporary machine learning models. The entire dataset was tokenized resulting in 285,622 PHTs (63\% with at least one hospital admission), comprising over 360 million tokens. ETHOS outperformed benchmark models in predicting hospital admissions, ICU admissions, and prolonged stays, achieving superior AUC scores. Its risk estimates were robust across demographic subgroups, with calibration curves confirming model reliability. The explainability module provided valuable insights into patient-specific risk factors. ARES, powered by ETHOS, advances predictive healthcare AI by delivering dynamic, real-time, personalized risk estimation with patient-specific explainability. Although our results are promising, the clinical impact remains uncertain. Demonstrating ARES’s true utility in real‐world settings will be the focus of our future work. We release the source code to facilitate future research\footnote{\url{https://github.com/ipolharvard/ethos-ares}}.

\end{abstract}

% keywords can be removed

\clearpage

\section{Introduction}

The United States allocates nearly 18 percent of its GDP to healthcare~\cite{CMS2024NHE}, yet Americans have shorter lifespans and poorer health than residents of other high-income nations. Among these countries, the U.S. not only has the lowest life expectancy but also the highest rates of preventable deaths~\cite{Gunja2023-pk}. Hospitals face mounting challenges managing patient influx and identifying individuals at risk for critical outcomes, including mortality, intensive care unit (ICU) admission, or prolonged hospital stays~\cite{Committee-on-the-Future-of-Emergency-Care-in-the-United-States-Health-System2007-ml}. Accurate prediction of critical clinical events is essential for enhancing patient care and optimizing the timely allocation of limited healthcare resources~\cite{Yang2016-ip}. Early identification of at-risk patients enables clinicians to prioritize interventions, anticipate potential escalations in care, and improve outcomes while simultaneously reducing costs~\cite{Horton2020-nz,Adams2022-bd}. However, current methodologies often fail to fully utilize the vast and complex data available in electronic health records (EHRs), a limitation that becomes particularly evident in emergency settings where time-sensitive decisions are critical~\cite{Edelson2024-rz,Gerry2020-ka,Winslow2022-yw,Escobar2020-jd,Cummings2023-yl}. Traditional scoring systems, such as the National Early Warning Score (NEWS)~\cite{Williams2022-ah} and the Modified Early Warning Score (MEWS)~\cite{Subbe2001-jx}, rely on static variables and predefined thresholds, constraining their ability to adapt to dynamic and multifaceted patient data. These approaches are further hindered by their reliance on specific cutoff points for data inclusion (e.g., triage, 24-hour windows), which can overlook valuable longitudinal patterns.

\begin{figure}
    \centering
    \includegraphics[width=\textwidth]{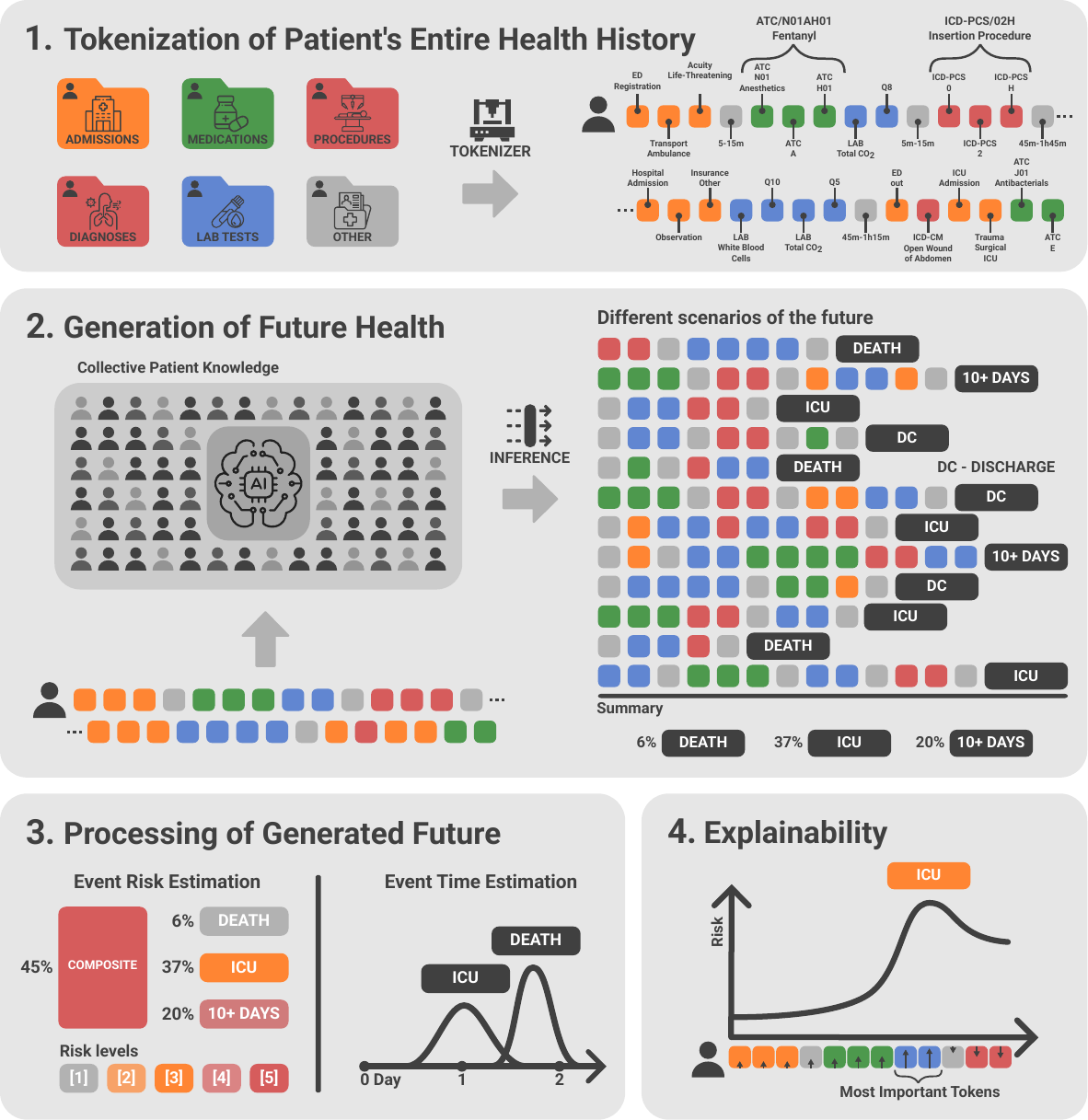}
    \caption{\textbf{Workflow of the Adaptable Risk Estimation Score (ARES) Framework.} This figure illustrates the ARES framework, developed on the ETHOS model, for dynamic and explainable risk evaluation. Panel 1 depicts the tokenization of a patient’s entire health history into structured events represented as a sequence of tokens (PHTs), incorporating standardized coding systems such as ATC for medications, ICD-PCS for procedures, and others. Panel 2 demonstrates how the ETHOS model trained on a large dataset of PHTs to simulate potential future patient health timelines (fPHTs). By analyzing a particular patient’s known PHT and generating multiple fPHTs, the model estimates the probabilities of critical outcomes, such as inpatient death, ICU admission, or a prolonged hospital stay exceeding 10 days. Panel 3 showcases the result of processing of fPHTs to calculate event-specific risks and predict the timing of these events, should they occur. Risk levels are defined across five categories, color-coded for enhanced clinical interpretability. Panel 4 showcases the explainability module, which identifies the key factors influencing specific risk estimates, offering personalized and actionable insights to support clinical decision-making. In this example, blue tokens indicate factors contributing to an increased risk of ICU admission.}
    \label{fig:workflow}
\end{figure}

Recent advances in generative machine learning—particularly transformer architectures~\cite{Vaswani2017-of,Renc2024-jf,Yang2023-ic,Li2023-lv} that underpin Large Language Models~\cite{Luo2024-bw,Thirunavukarasu2023-nh}—have unlocked unprecedented capabilities for processing high-dimensional, heterogeneous, time-stamped health data from EHRs~\cite{Kraljevic2024-fj,McDermott2023-zd,Steinberg2023-zx,Li2020-se,Jeong2023-hy}. In this work, we build on our Enhanced Transformer for Health Outcome Simulation (ETHOS)~\cite{Renc2024-jf}, which differs from prior efforts in its tokenization and handling of EHR events. ETHOS is autoregressively pretrained—without any task-specific labels—on over 321 million tokens drawn from 269,741 Patient Health Timelines (PHTs), learning broad, high-dimensional representations that transfer across tasks. Operating on PHTs (tokenized sequences of demographics, diagnoses, medications, see \autoref{tab:data-sources}), ETHOS generates plausible future timelines (\autoref{fig:workflow}) and delivers zero-shot predictions for mortality, ICU admission, prolonged stay, and composite endpoints without any additional fine-tuning. By virtue of its scale, generalizability, and multi-task adaptability, ETHOS serves as a {\em foundation model} for PHT generation.

\begin{figure}
    \centering
    \includegraphics[width=\textwidth]{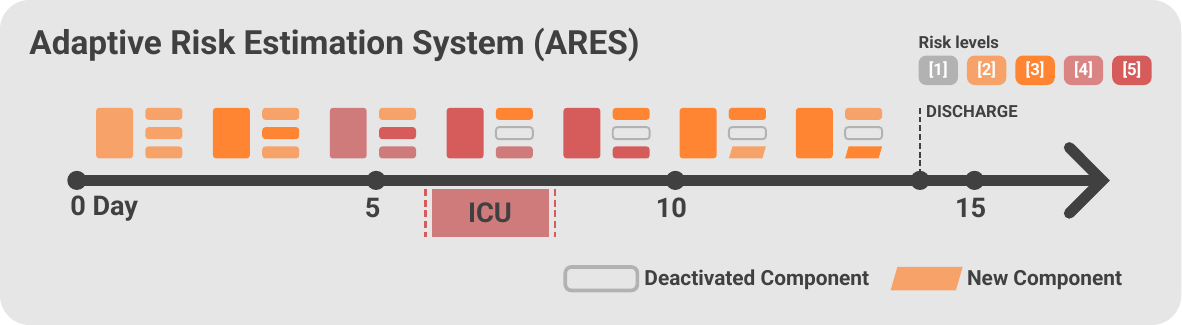}
    \caption{\textbf{Timeline of a Patient's Hospital Stay and Hypothetical Risk Predictions by ARES.} This figure illustrates the timeline of a patient’s hospital stay, from admission to discharge around Day 14, demonstrating how ARES dynamically adjusts its predictions based on the patient’s evolving clinical status and medical history. By Day 5, ARES predicts a high risk of ICU admission, which is subsequently confirmed as the patient is admitted around Day 6. Once the patient is in the ICU, ARES discontinues ICU risk evaluation, as indicated by the “Deactivated Component” label. After the ICU stay, ARES identifies an increased likelihood of a hospital stay exceeding 10 days. Upon reaching the 10-day threshold, ARES automatically recalibrates its predictions, replacing the previous risk estimation with the likelihood of a 15-day stay, now categorized as a “New Component” in the risk assessment.}
    \label{fig:ares}
\end{figure}

Once trained, ETHOS can generate multiple simulated future patient health timelines (PHTs) and estimate the probability of clinical events occurring within those trajectories (e.g., ICU admission). For adverse events during an inpatient stay, these probabilities serve as dynamic risk estimates, effectively functioning as an early warning system. Unlike traditional methods that require separate models or task-specific retraining, ETHOS operates as a unified model capable of concurrently assessing multiple clinical endpoints. As new patient data become available, risk estimates are automatically updated. This flexible and scalable risk prediction framework, built on ETHOS, is referred to as the Adaptive Risk Estimation System (ARES), as illustrated in \autoref{fig:ares}. Risk is quantified into five ordinal categories (levels 1 through 5) based on the predicted probability: 0–20\% corresponds to level 1, 20–40\% to level 2, and so on.

In this paper, we present ARES and introduce a novel explainability framework that delivers fully personalized insights, potentially allowing clinicians to understand the specific factors influencing the system’s risk predictions for individual patients. We benchmark the performance of ARES against state-of-the-art methods across multiple clinically relevant tasks, demonstrating its superior predictive accuracy. We validate its effectiveness and provide the accompanying code for the full reproduction of all the experiments by other researchers.

\section{Methods}

\subsection{ETHOS and Probabilistic Inference}

We introduced ETHOS in \cite{Renc2024-jf}. It operates on Patient Health Timelines (PHTs), which are tokenized chronological representations of patient medical histories (see~\autoref{tab:original_vs_tokenized}). Here, tokenization refers to encoding clinical events, such as inpatient visits, procedures, laboratory results, medication administrations, and vital signs, as sequences of discrete tokens.  The intervals between events are captured using specialized time‐interval tokens.  Formally, a PHT is a sequence of integer labels corresponding to these tokens, and its length can reach hundreds of thousands of tokens.

ETHOS employs a transformer‐based generative model to predict future clinical events from tokenized PHTs.  During inference, ETHOS generates successive tokens, each denoting a prospective future event, until a predefined stopping condition is met, such as the appearance of a target event token or the attainment of a simulation time limit.  By repeatedly simulating multiple future PHTs (fPHTs) for each patient, ETHOS explores a range of possible trajectories, thereby quantifying the inherent uncertainty in its predictions.  For instance, if \(N\) fPHTs are simulated and \(M\) of these trajectories include an inpatient mortality token, the estimated mortality probability is given by $M/N$ (see~\autoref{sec:monte-carlo-just}). All probabilistic inferences in this paper utilize Monte Carlo (MC) sampling with $N=100$ simulated fPHTs per patient, which inevitably introduces variability due to finite draws. We quantify this uncertainty by modeling the number of positive outcomes as a $Binomial(N,p)$ random variable and computing 95\% confidence intervals, and visualizing these as shaded bands around the mean risk trajectory (e.g., Figure~\ref{fig:ares-timeline}).

For detailed information on the transformer architecture, PHT statistics, and tokenization procedures, as well as intuitive explanation of ETHOS, please refer to our first publication~\cite{Renc2024-jf} and \autoref{sec:intuitive-ethos}.

\subsection{Data}

In this study, we used the Medical Information Mart for Intensive Care (MIMIC-IV) version 2.2 database~\cite{Johnson2023-cv,Johnson2023-ue}, including its ED extension. MIMIC-IV, developed by the Massachusetts Institute of Technology and Beth Israel Deaconess Medical Center contains de-identified health records for almost 300,000 patients either admitted to the ED and/or hospital at BIDMC from 2008 to 2019. Detailed patient demographics are presented in~\autoref{tab:population-demographics}.

We extracted relevant data from the MIMIC-IV tables as detailed in~\autoref{tab:data-sources}. Laboratory tests and medications were standardized using ATC codes, and all diagnostic and procedural codes were mapped to ICD-10 when necessary, as described in~\cite{Renc2024-jf}. Additional tables requiring advanced processing, such as clinical notes, were not included in the current implementation of ETHOS.

The dataset was split into two disjoint groups: training/validation (90\%) and testing (10\%). Exactly the same splits were used for all methods investigated.

\subsection{Tokenization, PHT Construction, model training}

The core of ETHOS lies in constructing PHTs from electronic medical records (EMRs) using a tokenization strategy that captures diverse clinical events. A PHT represents a patient's medical history as a sequence of tokens, each encoding specific health-related information organized chronologically. This structured representation enables comprehensive modeling of patient journeys and more accurate clinical predictions. To build PHTs, we used the MEDS-DEV~\cite{UnknownUnknown-ib} extraction pipeline that converts EHR data to an intermediate format called MEDS~\cite{Arnrich2024-rc} to facilitate further data transformations. Advanced transformation operations were subsequently applied, breaking down each event into 1 to 7 tokens based on its complexity. 

For example, lab test results were encoded using quantile-based tokens to represent clinical significance. Time-interval tokens were added to mark the elapsed time between successive events, with intervals shorter than 5 minutes omitted and longer gaps tokenized into 19 distinct interval tokens. Continuous numerical values, such as lab test results, were similarly quantile-encoded using ten quantiles, balancing clinical interpretability and predictive precision. Diagnostic and procedural codes, including ICD-10-CM, ICD-10-PCS, and ATC drug codes, were encoded hierarchically, which leveraged their inherent structure to enhance the transformer model's attention mechanisms. For more details, refer to~\cite{Renc2024-jf}.

Static patient attributes such as gender, marital status, race, and body mass index (BMI) were encoded using a single token depending on the value. For age, tokens of quantiles were reused, allowing age representation from 0 to 99. For instance, a 46-year-old patient would be coded as Q5 and Q7. Attributes with potential variability were represented using their most recently known value at the start of the timeline. By incorporating these elements, ETHOS ensured a richer and more adaptable representation of patient timelines.

During the training phase, 6 million tokens (1.8\% of the train/validation dataset) were used for validation to balance model optimization and computational efficiency. The detailed statistics about the tokenized dataset are available in~\autoref{tab:simple-pht-stats},~\ref{tab:detailed-token-stats}, and information about the model is in~\autoref{fig:model-info}. 

\subsection{Explainability}

\begin{figure}
    \centering
    \includegraphics[width=\textwidth]{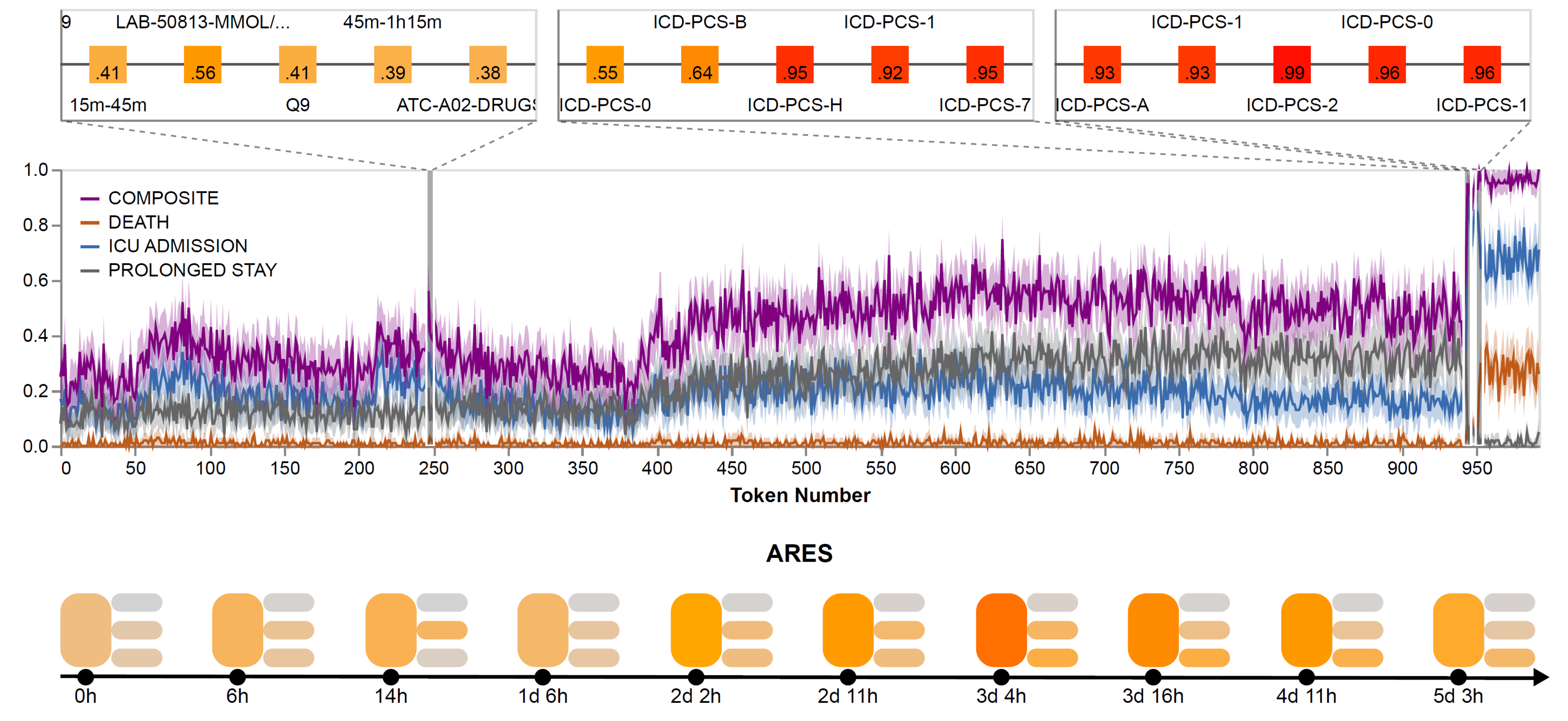}
    \caption{\textbf{ARES Risk Trajectories.} This figure illustrates risk trajectories for nearly 1000 tokens preceding patient death, as monitored by ARES, which evaluates the probability of death, ICU admission, prolonged hospital stay, and a composite risk score. The lower panel provides a color-coded representation of risk with the actual time since the ED presentation. In contrast, the upper panel highlights three 5-token regions influencing risk predictions at areas marked by the thin gray bar. In the first region, token LAB-50813 (Lactate Blood Test) increases the composite risk score from 0.41 to 0.56, but since the result falls in Q9 (80–90th percentile), ETHOS downgrades the risk estimate back to the previous level. In the second region (close to the end), a sharp increase in composite risk occurs due to heightened ICU admission triggered by ICD-PCS code 0BH17EZ, which is coded by 7 tokens (only 5 visible), which represents Endotracheal Airway Insertion into the Trachea via Natural or Artificial Opening. The 'H' token specifically signals ETHOS to escalate the ICU risk to nearly 1.0, indicating that the patient is being intubated de novo. The ICD-10-PCS breakdown confirms the procedure as a respiratory intervention involving tracheal insertion via a natural or artificial opening. ICD-PCS 0BH17EZ does not increase the risk of death, but the next ICD-PCS 5A12012 (5 tokens coding A1202 visible) raises the risk of death to about 0.25. We note that an increased risk of death is associated with a decreased risk of ICU admission, as these are competing risks. This visualization demonstrates how ARES dynamically adjusts risk scores based on evolving patient data, integrating clinical trajectories into real-time risk assessment. In this example, the rapid risk increases immediately following invasive procedures (e.g., intubation) should be interpreted as retrospective severity markers rather than actionable alerts, since they occur too late to guide effective intervention. Shaded bands around each trajectory denote the 95\% confidence intervals arising from Monte Carlo sampling.}
    \label{fig:ares-timeline}
\end{figure}

As illustrated in~\autoref{fig:ares-timeline}, stochastic simulations can be initiated not only from the most recent token representing current information but also from any preceding token in the patient’s history. This allows risk estimates to be visualized as a time series, highlighting how specific medical events affect risk over time. This approach provides intuitive visualizations, offering clinicians clear insights into the factors contributing to current risk values.

\subsection{Methods used for benchmarking}

We followed benchmarking tasks for emergency department models presented in the Emergency Department MIMIC-IV-ED benchmark paper~\cite{Xie2022-ur}. Three tasks were defined: prediction of the hospital admission at triage, prediction of the critical outcome (death or transfer to ICU within 12 hours) at triage, and ED re-presentation within 72 hours after discharge from ED. We applied machine learning methods (logistic regression, random forest, gradient boosting), scoring systems MEWS~\cite{Subbe2001-jx}, NEWS~\cite{Williams2022-ah,Smith2013-jz,Zhang2025-gp}, Rapid Emergency Medicine Scores (REMS)~\cite{Olsson2004-zw}, Cardiac Arrest Risk Triage (CART)~\cite{Churpek2012-ie}, five-level triage system Emergency Severity Index (ESI)~\cite{Eitel2003-nq} and neural networks-based models including multilayer perceptron, Med2Vec~\cite{Choi2016-fe} and Long Short-Term Memory (LSTM)~\cite{Hochreiter1997-af}.

\begin{figure*}[h]
    \centering
    \includegraphics[width=0.7\textwidth]{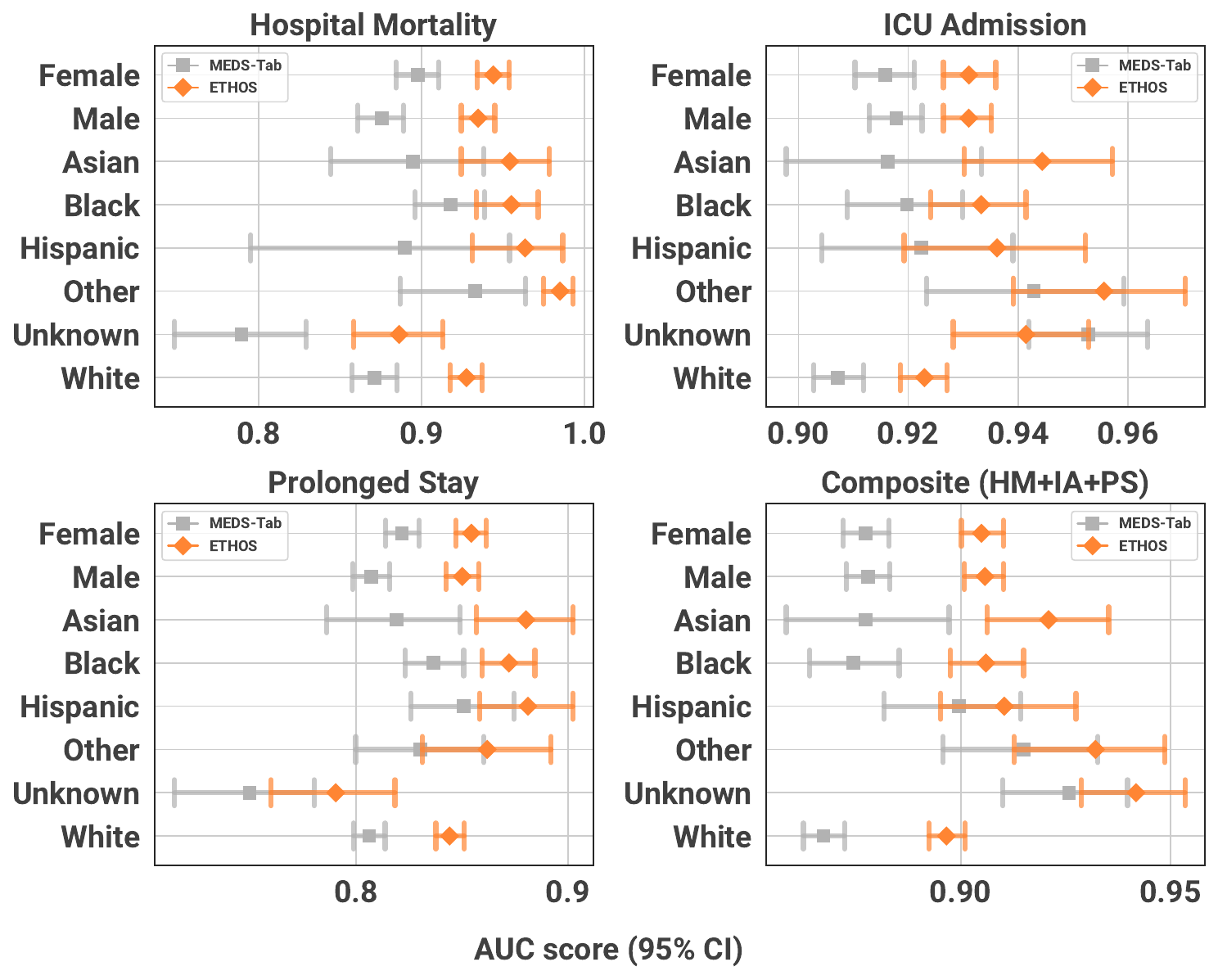}
    \caption{\textbf{AUC Comparison Between ETHOS and MEDS-Tab Across Demographic Subgroups and Prediction Tasks.} AUC scores with 95\% confidence intervals are shown for ETHOS (orange) and MEDS-Tab (gray) across four prediction tasks: Hospital Mortality, ICU Admission, Prolonged Stay, and Composite Outcome (Hospital Mortality + ICU Admission + Prolonged Stay). Performance is reported for the overall population and stratified by gender (Female, Male) and race (Asian, Black, Hispanic, Other, Unknown, White). ETHOS consistently outperforms MEDS-Tab across all demographic subgroups and tasks.}
    \label{fig:ares-results}
\end{figure*}

To compare tasks used for early warning scores, we compared the MEDS-Tab library~\cite{Oufattole2024-uy} which was used to establish a baseline. MEDS-Tab converts time-series EHR data into a tabular format by aggregating features across multiple time windows. It takes longitudinal patient data and applies various aggregation functions (like sum, count, min, max) over different historical window sizes to create fixed-size feature vectors, where each feature represents a combination of a medical code, time window, and aggregation method. XGBoost~\cite{Chen2016-tf} models are trained on these tabular features computed from data windows prior to each prediction time point for each clinical task.

\subsection{Statistical Methods}

The performance of predictive models was evaluated using Receiver Operating Characteristic (ROC) curves and corresponding Area Under the Curve (AUC) values. Bootstrapping techniques were employed to estimate 95\% confidence intervals (CIs) for AUCs. Model predicted probabilities were compared with observed event frequencies using calibration curves to evaluate ETHOS's reliability and alignment with real-world clinical outcomes. All statistical analyses were conducted using Python-based libraries, including scipy and scikit-learn~\cite{Virtanen2020-rd,Pedregosa2011-pl}. Data visualization, including ROC curves, calibration plots, and other statistical figures, was performed using matplotlib, seaborn and altair.

\section{Results}

Following the tokenization process, the data of 299,721 unique patients from the MIMIC-IV dataset was converted into 285,622 PHTs, which were subsequently used for training and testing. Patients were excluded if they had no usable data after tokenization. This occurs when patients in MIMIC have little or no structured information available, or when the available information (e.g., clinical notes, imaging) is not tokenized in the current ETHOS version. Of the total PHTs, approximately 63\% (180,733) contained hospital admissions records. The tokenized dataset comprised over 360 million tokens in total. In the Supplementary Materials, we provide  detailed information regarding the MIMIC-IV data used (\autoref{tab:data-sources}), patient demographics (\autoref{tab:population-demographics}), characteristics of the PHTs (\autoref{tab:simple-pht-stats}) and tokens (\autoref{tab:detailed-token-stats}). The model was trained and validated on 90\% of the PHTs, with the remaining 10\% reserved for testing. During inference, at least $N=100$ fPHTs were generated for each investigated task.

\begin{figure}[h]
    \centering
    \includegraphics[width=0.7\textwidth]{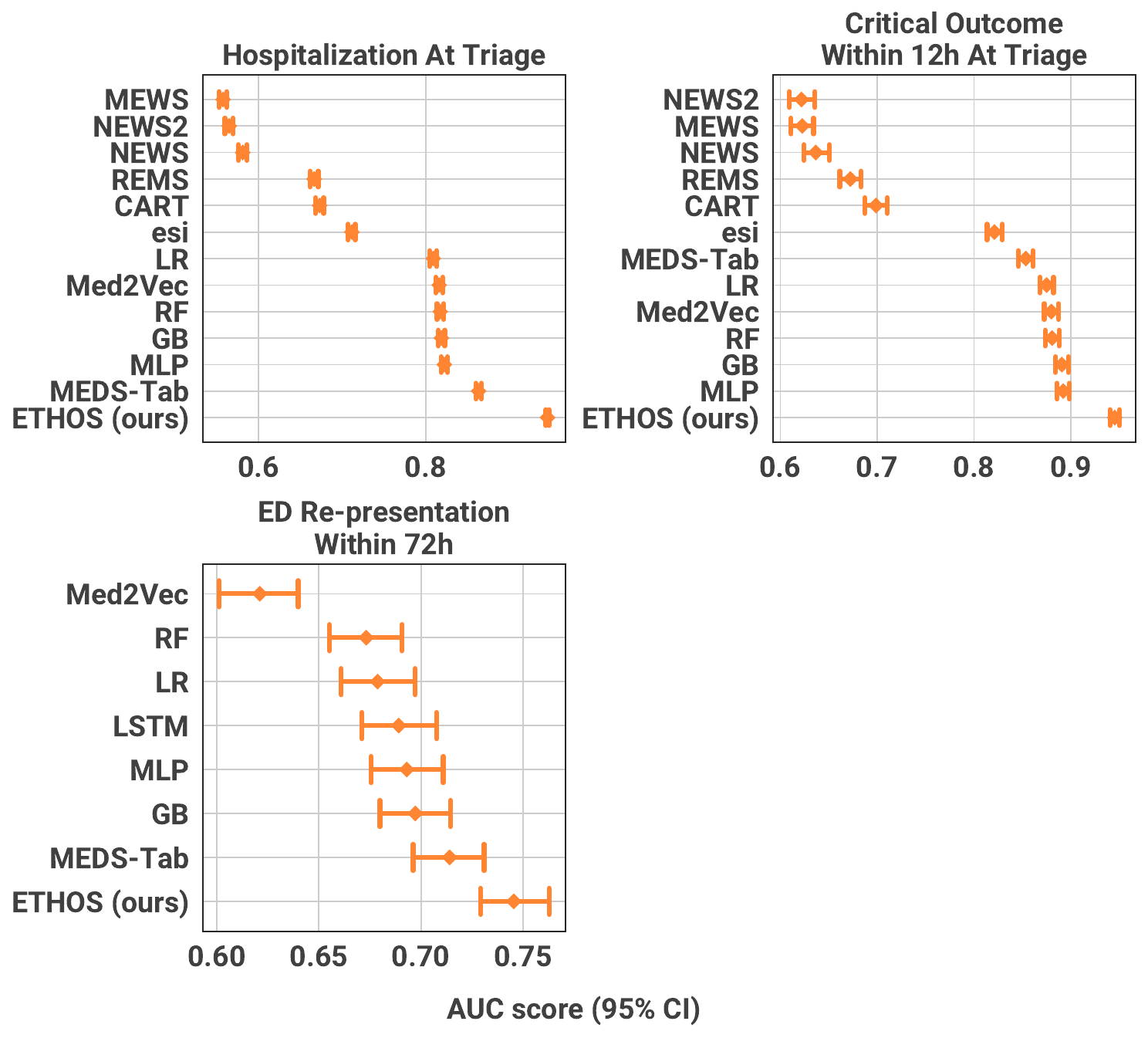}
    \caption{\textbf{Predictive results for the ED benchmark tasks.} Fewer methods appear in the ED re-presentation task (right) because score-based approaches, designed specifically to estimate in-hospital deterioration, are not applicable once the patient has left the ED. Ethos consistently achieves the best performance across all evaluated tasks.}
    \label{fig:ed-bench-results}
\end{figure}

The predictive performance of ARES and MEDS-Tab was evaluated on three individual clinical endpoints—hospital mortality (HM), ICU admission (IA), and prolonged hospital stay (PS; defined as length of stay >90th percentile)—and, as an illustrative demonstration of joint risk modeling, on a composite criterion combining these events (HM-IA-PS). The prevalence of these tasks is: 1.85\%, 15.44\%, 9.01\% and 20.39\%, respectively. This composite endpoint demonstrates ARES’s capacity to compute joint probabilities across heterogeneous outcomes and to naturally model their statistical dependencies. The composite score represents the cumulative risk of clinician-defined critical events. All predictions were generated at the hospital admission.
As summarized in~\autoref{fig:ed-bench-results}, \autoref{fig:ares-results}, and~\autoref{tab:ares-results}, ARES consistently outperformed MEDS-Tab across both individual and composite endpoints, achieving higher AUC values in every case.  Notably, these gains were observed across all racial subgroups, with the most pronounced improvements for Asian and Hispanic patients, indicating ARES’s robustness and its potential to reduce disparities in predictive accuracy.

\autoref{fig:ares-timeline} illustrates the dynamic risk trajectories generated by ARES, showcasing how the system continuously updates probability estimates for key clinical outcomes, including ICU admission, prolonged hospital stay, and mortality, as new clinical events occur. The figure highlights specific medical interventions, such as laboratory tests and procedures, that drive significant changes in risk estimates, demonstrating ARES’s ability to integrate evolving patient data into real-time risk assessment. The results underscore the model’s capacity to capture complex temporal relationships between clinical events, dynamically recalibrating risk scores based on patient status and treatment progression. 

In addition to risk which are part of ARES and to contextualize the predictive capabilities of ETHOS, we compared its performance against traditional early warning scores and other ML models. \autoref{fig:ed-bench-results} presents the AUC values (ROC curves in \autoref{fig:ethos-auroc}) for key ED benchmark tasks: hospitalization at triage, critical outcomes within 12 hours of triage, and ED re-presentation within 72 hours post-discharge. ETHOS demonstrated consistently superior predictive accuracy across all evaluated tasks. We provide detailed numerical values in the Supplement (\autoref{tab:ed-hospitalization},\ref{tab:ed-critical-outcome},\ref{tab:ed-representation}).

The risks provided by ETHOS were also found to be well-calibrated, as tested by calibration curves. Brier scores were found in the range 0.01-0.14 depending on the task, indicating excellent to good performances, as shown in~\autoref{fig:ethos-calibration}.

\section{Discussion}

The ARES framework introduces an innovative approach to building predictive models by leveraging cutting-edge artificial intelligence technology. Several aspects of this approach distinguish it from traditional models. First, ARES enables dynamic risk estimation at any time during a patient’s stay, from admission to discharge. Powered by ETHOS~\cite{Renc2024-jf}, ARES utilizes PHTs and incorporates all available clinical information at the time of risk estimation. Unlike traditional models, which rely on static data points such as information collected within 24 hours after admission or ED presentation or data up to triage~\cite{Xie2022-ur,Meng2022-bk}, ARES continuously adapts to the patient’s evolving clinical status. This adaptability overcomes a key limitation of static models, which may not perform optimally outside the narrow time frames for which they are designed. This capability is demonstrated in the accompanying~\autoref{fig:ares-timeline} and~\autoref{tab:timeline-examples}, which illustrate how risk evolves over time during a patient’s hospital stay. These visualizations, which depict how personalized risk evolves over time to reach the current estimates, provide insights into the specific factors driving model predictions for each patient. They highlight clinical events associated with increased or decreased risk, offering real-time explainability. By identifying the most influential features contributing to an individual’s risk assessment, ARES have the potential to empower clinicians with a clearer understanding of the rationale behind each prediction. 

As illustrated in~\autoref{fig:workflow}, ARES can estimate risk for various critical events, such as in-hospital mortality, ICU admission, and prolonged hospital stays. Beyond these standard metrics, additional indicators can be integrated seamlessly, including the risk of ICU admission during a specific length of stay, ICU readmission, acute kidney injury, sepsis, cardiac arrest, or 30-day readmission, and others. The ETHOS model, which underpins ARES, allows for the dynamic combination of these risks into composite measures while accounting for their interdependencies. For example, the occurrence of mortality on day 8 would render the probability of a 10-day hospital stay zero. This ability to incorporate conditional and causal relationships between tracked events is another strength of ARES. Importantly, integrating additional metrics does not require model retraining or modifications of ETHOS. Once a range of possible future PHTs has been generated, any additional metrics can be calculated with minimal computational resources, making ARES scalable and adaptable to diverse healthcare settings.

In its current implementation, ETHOS distills multiple fPHTs into a single predictive decision, such as inpatient mortality. However, this approach overlooks the wealth of longitudinal information contained in these trajectories, including the sequence of clinical events that lead to a particular outcome, or the absence thereof. By merely predicting the likelihood of an adverse event, valuable insights into the pathways that contribute to deterioration or recovery remain underutilized. Expanding ARES to provide a more granular, trajectory-based interpretation of risk would allow clinicians not only to assess a patient’s probability of experiencing a critical event but also to understand the evolving clinical course leading to that outcome including the cost. This enhanced approach would address a key limitation highlighted in the early warning paradox~\cite{Logan-Ellis2025-qr}, where models trained on retrospective data may fail to capture the full complexity of clinical interventions and their effects on patient outcomes. Moving forward, we aim to refine ARES to incorporate and visualize these probabilistic trajectories. This will equip clinicians with deeper, more actionable insights into clinical risk dynamics and potentially provide new information about causality in patient outcomes.

We recognize that, to date, ARES has not undergone formal usability testing with frontline clinicians, yet their ultimate impact depends on seamless integration into real‐world workflows. Emergency medicine specialists on our team have provided informal feedback on feasibility and clarity of the risk estimates and explanatory highlights, and we are now designing pilot simulations in which physicians will “round” on de‐identified patient cases presented through mock electronic charts powered by ARES. These studies, first leveraging MIMIC‐derived timelines and subsequently our own Mass General Brigham data, will allow us to observe decision points, gather qualitative feedback on timing and interpretability of alerts, and refine both the user interface and explanation formats. We anticipate that iterative, case‐based testing will guide the development of a clinician‐centered dashboard, ensuring that ARES’s predictions align with care priorities and support timely, actionable insights in the emergency setting.

This study has several important limitations. First, although we demonstrated ETHOS using PHTs derived from the MIMIC-IV-ED dataset, its performance on data from other institutions may be compromised without retraining on external cohorts. Electronic health record systems and clinical workflows differ substantially across hospitals—driven by variations in documentation practices, patient case‐mix, and care protocols—so models trained on one site can yield misleading risk estimates when deployed elsewhere. Moreover, our training data may harbor demographic and institutional biases (for example, overrepresentation of certain age, race, or socioeconomic groups), which could impair generalizability and exacerbate health inequities if unaddressed. The MIMIC dataset is relatively small, and although it contains dense, diverse information, the limited cohort size inevitably constrains generalization. Our model is explicitly designed to scale, and we expect its performance to further improve when trained on larger and more diverse datasets that capture a broader range of clinical variability. We have not yet conducted a thorough fairness audit to quantify potential disparities in ETHOS’s predictions across sex, race, or ethnicity. By contrast, in domains such as radiology or pathology, data inputs like images are relatively standardized, enabling easier cross-institution transfer. To facilitate broader validation and retraining, we have ensured that the ETHOS-ARES codebase is fully compatible with the MEDS health AI data standard~\cite{Arnrich2024-rc}. This interoperability simplifies the process for other researchers to apply identical model architectures to their local data, perform bias and subgroup analyses, and iteratively refine ETHOS for diverse patient populations.

We also recognize that our evaluation on the extensively curated MIMIC-IV dataset may underestimate challenges encountered in real-world EHRs, which often exhibit higher rates of missing or irregular data, temporal shifts in documentation and care processes, and evolving patient populations. Although ETHOS is designed to operate on incomplete timelines, elevated missingness will still impair model accuracy. It is unclear if temporal biases or practice changes may introduce drift over time for ETHOS because the inferences are based on the context which contemporary to predictions, but this has not been investigated yet. Future work will systematically assess ETHOS’s resilience to these factors and develop strategies for ongoing recalibration in heterogeneous clinical environments. In addition, our current benchmarking employed only classical risk-prediction methods. While sufficient to demonstrate ETHOS’s competitiveness on established tasks, more sophisticated approaches could yield higher benchmark scores. Even if such models were to match or slightly exceed our performance on static benchmarks, this would not diminish ETHOS’s primary contribution: a dynamic, explainable framework that adapts predictions in real time as new clinical data become available. Future work will expand benchmarking to include a broader set of advanced methods (e.g.~\cite{Steinberg2023-zx}).

In current implementation, we exclude unstructured clinical text and because of that ETHOS may miss nuanced patient information, such as narrative impressions or social determinants, that could enhance risk estimation and zero‐shot generalizability. Integrating free-text notes poses challenges in segmenting and embedding variable-length narratives alongside structured events without overwhelming the model’s capacity. In future work, we will explore the use of pretrained clinical-language-model embeddings, hierarchical chunking of note content, and multimodal fusion techniques to incorporate these rich data into the PHTs.

Data standardization is often proposed as a solution to address the challenges of variability in healthcare data. However, achieving meaningful standardization would require identifying commonalities between healthcare systems, an endeavor that may not be feasible given the diversity of clinical practices, patient populations, and institutional workflows. An alternative is to train AI models, such as ETHOS, on raw data from diverse institutions, allowing the model itself to learn and interpret the underlying patterns and clinical pathways. This approach mirrors the capability of large language models to discern meaning from vastly different styles of text and presentations or even different languages, leveraging the same transformer architecture as ETHOS. We performed an energy consumption analysis comparing training of ETHOS to other known LLMs that can be seen in \autoref{tab:energy-comparison}.

In summary, recent advances in AI have created unprecedented opportunities for innovative solutions like ARES, which harness large volumes of heterogeneous data to build general‐purpose models whose predictive performance exceeds state‐of‐the‐art methods. ARES delivers dynamic, personalized risk estimates and offers real‐time explainability, empowering clinicians to make better‐informed decisions. Moreover, its modular architecture and the underlying ETHOS framework enable seamless integration of additional data modalities, such as radiology, genomics, and other institutional datasets, further enhancing predictive accuracy and broadening applicability across diverse healthcare environments. Although our results are promising, the clinical impact remains uncertain. Demonstrating ARES’s true utility in real‐world settings will be the focus of our future work.  

As healthcare costs and complexity continue to rise, PHT-based frameworks like ARES show a promising pathway towards data-driven AI-enabled individualized patient care with the potential to reduce morbidity, improve outcomes, and lower healthcare costs.

\section*{Acknowledgments}
We thank Kinga Renc, M.Arch, for her invaluable assistance with graphic design. We also thank Ethan Steinberg for the careful and very helpful review of our codebase. This work was supported in part by National Institutes of Health (NIH) grant number HL159183. 

\printbibliography

\clearpage
\appendix

\renewcommand{\thetable}{S\arabic{table}}
\renewcommand{\thefigure}{S\arabic{figure}}
\setcounter{figure}{0}

\section*{Supplementary Materials}
\label{sec:supplementary}

\begin{figure}[h]
    \centering
    \caption{\textbf{Model Architecture and Hyperparameter Overview.} (Left) The architecture of the transformer-based model, following the standard GPT design, includes multiple layers of masked multi-head attention and feed-forward modules, normalized at each step and combined with positional encodings. (Right) Summary of the hyperparameters used for model training and their explored ranges. The final model uses 6 layers, a context size of 2048, an embedding size of 768, 12 attention heads, a dropout rate of 0.3, and a batch size of 32. Additional information includes the percentage of discarded ambiguous inference repetitions (0.2–0.3\%) that appear when doing zero-shot inference.}
    \includegraphics[width=0.7\textwidth]{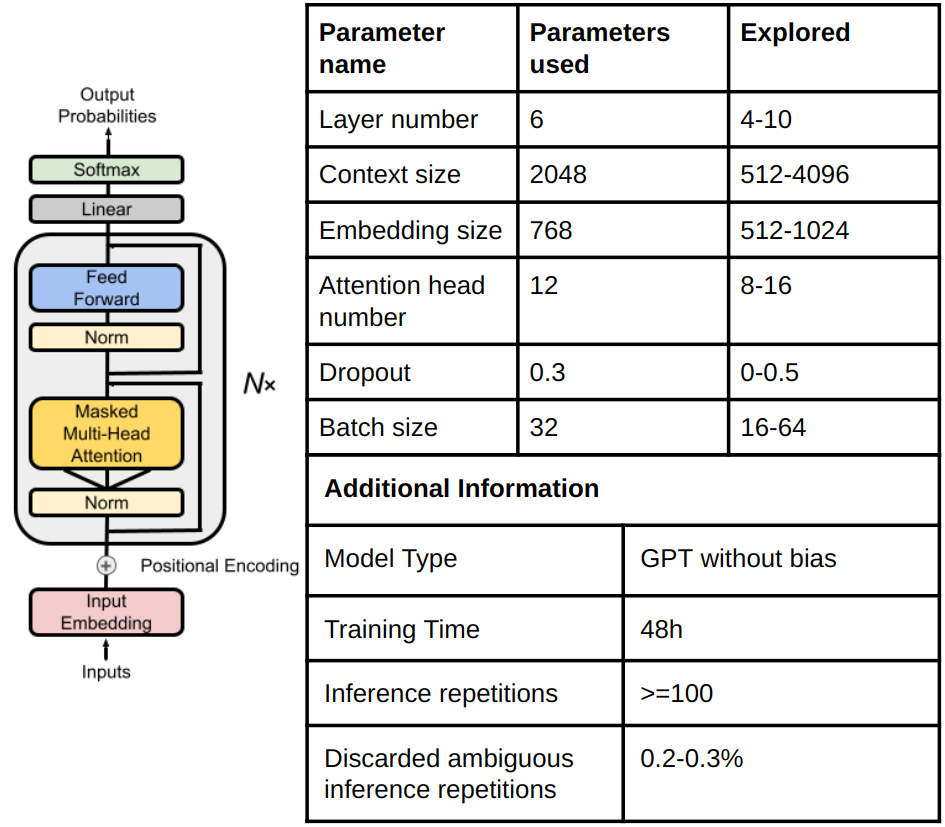}
    \label{fig:model-info}
\end{figure}

\begin{table}%
\centering%
\caption{\textbf{ETHOS performance on ARES tasks with a breakdown for demographic subgroups.} This table presents the predictive performance (AUROC with 95\% confidence intervals) of ETHOS (top) and MEDS-Tab (bottom) for four critical clinical outcomes used in ARES: Hospital Mortality, ICU Admission, Prolonged Hospital Stay (>10 days), and a Composite Risk Score (HM+IA+PS). The prevalence rates of each outcome are provided for reference. Performance metrics are further stratified by gender and race to assess potential disparities in model performance across demographic subgroups.}%
\label{tab:ares-results}%
\begin{tabular}{lcccc}%
\toprule%
&\textbf{Hospital Mortality}&\textbf{ICU Admission}&\textbf{Prolonged Stay}&\textbf{Composite (HM+IA+PS)}\\%
\textit{Prevalence} (\%)&1.95&15.44&9.01&20.41\\%
\toprule%
\multicolumn{5}{c}{\textbf{\Large{ETHOS}}}\\%
\toprule%
\textbf{Overall}&0.940 {[}0.932, 0.947{]}&0.932 {[}0.928, 0.935{]}&0.853 {[}0.848, 0.858{]}&0.906 {[}0.902, 0.909{]}\\%
\midrule%
\textbf{Gender}&&&&\\%
\hspace{1em} Female&0.944 {[}0.933, 0.953{]}&0.931 {[}0.927, 0.936{]}&0.854 {[}0.847, 0.862{]}&0.905 {[}0.900, 0.910{]}\\%
\hspace{1em} Male&0.935 {[}0.924, 0.945{]}&0.931 {[}0.926, 0.936{]}&0.850 {[}0.842, 0.857{]}&0.906 {[}0.901, 0.911{]}\\%
\midrule%
\textbf{Race}&&&&\\%
\hspace{1em} Asian&0.954 {[}0.925, 0.978{]}&0.944 {[}0.930, 0.958{]}&0.880 {[}0.858, 0.900{]}&0.921 {[}0.905, 0.936{]}\\%
\hspace{1em} Black&0.955 {[}0.935, 0.972{]}&0.933 {[}0.924, 0.943{]}&0.872 {[}0.858, 0.885{]}&0.906 {[}0.897, 0.915{]}\\%
\hspace{1em} Hispanic&0.964 {[}0.929, 0.987{]}&0.936 {[}0.918, 0.951{]}&0.881 {[}0.856, 0.903{]}&0.910 {[}0.893, 0.927{]}\\%
\hspace{1em} Other&0.985 {[}0.975, 0.993{]}&0.956 {[}0.937, 0.972{]}&0.862 {[}0.831, 0.889{]}&0.932 {[}0.915, 0.949{]}\\%
\hspace{1em} Unknown&0.886 {[}0.852, 0.911{]}&0.941 {[}0.928, 0.953{]}&0.790 {[}0.757, 0.819{]}&0.942 {[}0.928, 0.954{]}\\%
\hspace{1em} White&0.928 {[}0.918, 0.937{]}&0.923 {[}0.919, 0.927{]}&0.844 {[}0.838, 0.851{]}&0.897 {[}0.892, 0.901{]}\\%
\toprule%
\multicolumn{5}{c}{\textbf{\Large{MEDS-Tab}}}\\%
\toprule%
\textbf{Overall}&0.887 {[}0.877, 0.897{]}&0.918 {[}0.914, 0.921{]}&0.815 {[}0.810, 0.821{]}&0.879 {[}0.875, 0.883{]}\\%
\midrule%
\textbf{Gender}&&&&\\%
\hspace{1em} Female&0.898 {[}0.884, 0.910{]}&0.916 {[}0.910, 0.921{]}&0.822 {[}0.814, 0.830{]}&0.877 {[}0.872, 0.883{]}\\%
\hspace{1em} Male&0.876 {[}0.861, 0.889{]}&0.918 {[}0.913, 0.922{]}&0.807 {[}0.798, 0.816{]}&0.878 {[}0.873, 0.883{]}\\%
\midrule%
\textbf{Race}&&&&\\%
\hspace{1em} Asian&0.895 {[}0.844, 0.938{]}&0.916 {[}0.898, 0.933{]}&0.819 {[}0.786, 0.849{]}&0.877 {[}0.858, 0.897{]}\\%
\hspace{1em} Black&0.918 {[}0.896, 0.938{]}&0.920 {[}0.909, 0.930{]}&0.836 {[}0.823, 0.851{]}&0.874 {[}0.864, 0.885{]}\\%
\hspace{1em} Hispanic&0.890 {[}0.795, 0.954{]}&0.922 {[}0.904, 0.939{]}&0.851 {[}0.826, 0.874{]}&0.900 {[}0.882, 0.914{]}\\%
\hspace{1em} Other&0.933 {[}0.887, 0.964{]}&0.943 {[}0.923, 0.959{]}&0.830 {[}0.800, 0.860{]}&0.915 {[}0.896, 0.933{]}\\%
\hspace{1em} Unknown&0.789 {[}0.748, 0.829{]}&0.953 {[}0.942, 0.964{]}&0.750 {[}0.714, 0.780{]}&0.926 {[}0.910, 0.940{]}\\%
\hspace{1em} White&0.871 {[}0.857, 0.885{]}&0.907 {[}0.903, 0.912{]}&0.806 {[}0.799, 0.814{]}&0.867 {[}0.862, 0.872{]}\\%
\bottomrule%
\end{tabular}%
\end{table}

\begin{table}%
\centering%
\caption{\textbf{Demographic characteristics of the dataset analyzed in this study.} The table summarizes key demographic attributes of the dataset, stratified into Train/Validation, Test, and Total splits. Patient numbers, mean age (with standard deviation), and distribution across gender, race, and marital status are shown, with percentages provided in parentheses. The data highlights the representation of each subgroup within the splits, providing context for the population characteristics in the dataset.}%
\label{tab:population-demographics}%
\begin{tabular}{lrrr}%
\toprule%
&\textbf{Train/Validation}&\textbf{Test}&\textbf{Total}\\%
\toprule%
\textbf{Patient Number}&269,741&29,971&299,712\\%
\textbf{Mean Age (Std.)}&48.5 (20.9)&48.6 (20.9)&48.5 (20.9)\\%
\midrule%
\textbf{Gender (\%)}&&&\\%
\hspace{1em} Female&142,696 (52.9)&15,857 (52.9)&158,553 (52.9)\\%
\hspace{1em} Male&127,045 (47.1)&14,114 (47.1)&141,159 (47.1)\\%
\midrule%
\textbf{Race (\%)}&&&\\%
\hspace{1em} Unknown&115,437 (42.8)&12,684 (42.3)&128,121 (42.7)\\%
\hspace{1em} White&110,408 (40.9)&12,369 (41.3)&122,777 (41.0)\\%
\hspace{1em} Black&21,410 (7.9)&2,321 (7.7)&23,731 (7.9)\\%
\hspace{1em} Hispanic&9,214 (3.4)&1,023 (3.4)&10,237 (3.4)\\%
\hspace{1em} Asian&6,802 (2.5)&787 (2.6)&7,589 (2.5)\\%
\hspace{1em} Other&6,470 (2.4)&787 (2.6)&7,257 (2.4)\\%
\midrule%
\textbf{Marital Status (\%)}&&&\\%
\hspace{1em} Unknown&114,234 (42.3)&12,603 (42.1)&126,837 (42.3)\\%
\hspace{1em} Married&70,269 (26.1)&7,811 (26.1)&78,080 (26.1)\\%
\hspace{1em} Single&60,915 (22.6)&6,793 (22.7)&67,708 (22.6)\\%
\hspace{1em} Widowed&14,243 (5.3)&1,670 (5.6)&15,913 (5.3)\\%
\hspace{1em} Divorced&10,080 (3.7)&1,094 (3.7)&11,174 (3.7)\\%
\bottomrule%
\end{tabular}%
\end{table}

\begin{table}
\centering
\caption{\textbf{Prediction of Hospitalization At Triage.} Performance comparison of various models for predicting hospitalization at triage, evaluated using AUROC, AUPRC, sensitivity, and specificity (95\% confidence intervals in brackets). The thresholds for sensitivity and specificity were determined by finding the operating point on the ROC curve closest to (0,1). ETHOS demonstrates superior performance across all metrics, outperforming all other methods, including traditional scoring systems and machine learning models.}
\label{tab:ed-hospitalization}
\begin{tabular}{lcccc}
\toprule
  & AUROC & AUPRC & Sensitivity & Specificity \\
\midrule
LR & 0.809 [0.805, 0.813] & 0.775 [0.769, 0.781] & 0.734 [0.721, 0.750] & 0.736 [0.721, 0.749] \\
Med2Vec & 0.816 [0.812, 0.820] & 0.782 [0.775, 0.788] & 0.751 [0.734, 0.770] & 0.728 [0.711, 0.745] \\
RF & 0.817 [0.814, 0.821] & 0.785 [0.779, 0.791] & 0.759 [0.736, 0.764] & 0.726 [0.720, 0.747] \\
GB & 0.819 [0.815, 0.823] & 0.792 [0.786, 0.798] & 0.753 [0.731, 0.770] & 0.728 [0.715, 0.751] \\
MLP & 0.822 [0.818, 0.826] & 0.796 [0.790, 0.802] & 0.754 [0.743, 0.775] & 0.734 [0.716, 0.745] \\
esi & 0.712 [0.707, 0.716] & 0.632 [0.625, 0.638] & 0.584 [0.577, 0.590] & 0.784 [0.779, 0.789] \\
NEWS & 0.581 [0.576, 0.586] & 0.555 [0.548, 0.561] & 0.563 [0.556, 0.569] & 0.546 [0.540, 0.553] \\
NEWS2 & 0.565 [0.560, 0.570] & 0.538 [0.532, 0.544] & 0.519 [0.512, 0.526] & 0.570 [0.564, 0.577] \\
REMS & 0.666 [0.661, 0.671] & 0.605 [0.598, 0.612] & 0.605 [0.552, 0.722] & 0.641 [0.545, 0.711] \\
MEWS & 0.558 [0.553, 0.562] & 0.521 [0.515, 0.527] & 0.296 [0.289, 0.302] & 0.812 [0.806, 0.817] \\
CART & 0.673 [0.668, 0.678] & 0.617 [0.610, 0.624] & 0.703 [0.696, 0.709] & 0.578 [0.571, 0.585] \\
MEDS-Tab & 0.863 [0.860, 0.866] & 0.879 [0.876, 0.883] & 0.746 [0.735, 0.754] & 0.820 [0.809, 0.835] \\
ETHOS (ours) & 0.946 [0.944, 0.947] & 0.945 [0.943, 0.947] & 0.868 [0.859, 0.876] & 0.864 [0.856, 0.873] \\
\bottomrule
\end{tabular}
\end{table}

\begin{table}
\centering
\caption{\textbf{Prediction of Critical Outcome Within 12h At Triage.} Performance comparison of various models for predicting critical outcomes within 12 hours of triage, evaluated using AUROC, AUPRC, sensitivity, and specificity (95\% confidence intervals in brackets). The thresholds for sensitivity and specificity were determined by finding the operating point on the ROC curve closest to (0,1). ETHOS achieves the highest performance across most of the metrics, substantially outperforming all other methods, including traditional scoring systems and machine learning models.}
\label{tab:ed-critical-outcome}
\begin{tabular}{lcccc}
\toprule
  & AUROC & AUPRC & Sensitivity & Specificity \\
\midrule
LR & 0.875 [0.868, 0.882] & 0.308 [0.288, 0.328] & 0.813 [0.792, 0.836] & 0.782 [0.766, 0.803] \\
Med2Vec & 0.880 [0.872, 0.887] & 0.324 [0.305, 0.346] & 0.817 [0.799, 0.852] & 0.787 [0.762, 0.804] \\
RF & 0.881 [0.873, 0.888] & 0.362 [0.343, 0.386] & 0.812 [0.794, 0.829] & 0.792 [0.788, 0.796] \\
GB & 0.891 [0.884, 0.897] & 0.389 [0.367, 0.412] & 0.836 [0.804, 0.848] & 0.788 [0.779, 0.812] \\
MLP & 0.892 [0.886, 0.898] & 0.372 [0.352, 0.396] & 0.845 [0.806, 0.855] & 0.784 [0.780, 0.823] \\
esi & 0.821 [0.814, 0.829] & 0.190 [0.178, 0.201] & 0.900 [0.887, 0.913] & 0.637 [0.632, 0.642] \\
NEWS & 0.637 [0.624, 0.651] & 0.139 [0.127, 0.154] & 0.461 [0.440, 0.483] & 0.796 [0.792, 0.801] \\
NEWS2 & 0.622 [0.610, 0.636] & 0.130 [0.118, 0.144] & 0.416 [0.402, 0.605] & 0.822 [0.533, 0.826] \\
REMS & 0.672 [0.661, 0.683] & 0.093 [0.087, 0.102] & 0.662 [0.642, 0.683] & 0.602 [0.597, 0.607] \\
MEWS & 0.623 [0.611, 0.634] & 0.101 [0.093, 0.110] & 0.445 [0.424, 0.466] & 0.772 [0.768, 0.776] \\
CART & 0.699 [0.687, 0.710] & 0.134 [0.123, 0.147] & 0.579 [0.557, 0.600] & 0.720 [0.716, 0.725] \\
MEDS-Tab & 0.853 [0.846, 0.861] & 0.513 [0.493, 0.531] & 0.735 [0.717, 0.752] & 0.764 [0.759, 0.771] \\
ETHOS (ours) & 0.945 [0.941, 0.950] & 0.696 [0.678, 0.712] & 0.876 [0.860, 0.898] & 0.873 [0.852, 0.889] \\
\bottomrule
\end{tabular}
\end{table}

\begin{table}
\centering
\caption{\textbf{Prediction of Emergency Department Re{-}presentation Within 72h.} Performance comparison of various models for predicting emergency department re-presentation within 72 hours, evaluated using AUROC, AUPRC, sensitivity, and specificity (95\% confidence intervals in brackets). The thresholds for sensitivity and specificity were determined by finding the operating point on the ROC curve closest to (0,1). ETHOS demonstrates superior performance, outperforming all other methods and showcasing its effectiveness for this challenging task.}
\label{tab:ed-representation}
\begin{tabular}{lcccc}
\toprule
  & AUROC & AUPRC & Sensitivity & Specificity \\
\midrule
LR & 0.679 [0.661, 0.697] & 0.161 [0.141, 0.185] & 0.562 [0.544, 0.645] & 0.699 [0.613, 0.718] \\
Med2Vec & 0.621 [0.601, 0.640] & 0.128 [0.110, 0.148] & 0.560 [0.477, 0.595] & 0.615 [0.568, 0.725] \\
RF & 0.673 [0.655, 0.691] & 0.150 [0.131, 0.173] & 0.642 [0.549, 0.665] & 0.599 [0.594, 0.693] \\
GB & 0.697 [0.680, 0.714] & 0.165 [0.143, 0.188] & 0.623 [0.592, 0.704] & 0.662 [0.582, 0.690] \\
MLP & 0.693 [0.675, 0.711] & 0.168 [0.147, 0.192] & 0.603 [0.579, 0.676] & 0.675 [0.607, 0.701] \\
LSTM & 0.689 [0.671, 0.708] & 0.164 [0.143, 0.186] & 0.595 [0.566, 0.653] & 0.680 [0.633, 0.722] \\
MEDS-Tab & 0.714 [0.696, 0.731] & 0.189 [0.167, 0.214] & 0.645 [0.575, 0.689] & 0.657 [0.617, 0.742] \\
ETHOS (ours) & 0.745 [0.728, 0.762] & 0.214 [0.190, 0.239] & 0.669 [0.611, 0.698] & 0.685 [0.657, 0.757] \\
\bottomrule
\end{tabular}
\end{table}

\begin{table}%
\centering%
\caption{\textbf{Summary of Token and Timeline Statistics.} This table presents a comprehensive overview of the token and timeline data in the training, test, and combined datasets. Key metrics include the total number of tokens and timelines, along with statistics on timeline lengths such as the longest timeline, median, mean, and shortest timeline. The number of unique timeline tokens is also reported. The final section breaks down the encoding of timeline tokens into categories, such as time intervals, quantiles, medications, diagnoses, procedures, laboratory results, vitals, and other clinical features. This summary highlights the diversity and complexity of the tokenized data used in the study.}%
\label{tab:simple-pht-stats}%
\begin{tabular}{lrrr}%
\toprule%
&\textbf{Train/Validation}&\textbf{Test}&\textbf{Total}\\%
\toprule%
\textbf{Tokens}&324,667,250&36,325,697&360,992,947\\%
\midrule%
\textbf{Timelines}&257,081&28,539&285,620\\%
\midrule%
\textbf{Timeline Lengths}\\%
\hspace{1em} Longest&221,346&107,147&221,346\\%
\hspace{1em} Q3&1,053&1,063&1,055\\%
\hspace{1em} Median&331&340&332\\%
\hspace{1em} Mean&1,262&1,272&1,263\\%
\hspace{1em} Q1&122&123&122\\%
\hspace{1em} Shortest&2&2&2\\%
\hspace{1em} Unique&13,244&4,989&13,791\\%
\midrule%
\textbf{Unique Timeline Tokens}&4,495&3,947&4,495\\%
\midrule%
\textbf{Timeline Tokens Encoding}\\%
\hspace{1em} Time Intervals&19&19&19\\%
\hspace{1em} Quantiles&10&10&10\\%
\hspace{1em} Medications&312&275&312\\%
\hspace{1em} Diagnoses&2,989&2,542&2,989\\%
\hspace{1em} Procedures&34&34&34\\%
\hspace{1em} Labs&200&200&200\\%
\hspace{1em} Vitals&6&6&6\\%
\hspace{1em} HCPCS&66&37&66\\%
\hspace{1em} Inpatient Stays&29&29&29\\%
\hspace{1em} Emergency Department&6&6&6\\%
\hspace{1em} DRGs&772&737&772\\%
\hspace{1em} BMI&10&10&10\\%
\bottomrule%
\end{tabular}%
\end{table}

\begin{table}%
\centering%
\caption{\textbf{Overview of the data sources and their corresponding columns used in this work from the MIMIC-IV database and its extension MIMIC-IV-ED.} The table groups the data into three main categories: ED (Emergency Department), hosp (Hospital), and ICU (Intensive Care Unit). For each category, the associated tables and the specific columns extracted for the study are listed, highlighting key variables relevant to patient care and outcomes, such as identifiers (e.g., stay\_id, hadm\_id), timestamps (e.g., intime, charttime), and clinical observations (e.g., vitalsign, labresults). These selections were guided by the objectives of the study to comprehensively model patient trajectories and outcomes.}%
\label{tab:data-sources}%
\begin{tabular}{ll}%
\toprule%
Data Source&Used Columns\\%
\toprule%
\textbf{ed}&\\%
\hspace{1em} diagnosis&\makecell[tl]{icd\_code, icd\_version, stay\_id}\\%
\vspace{0.5em}%
\hspace{1em} edstays&\makecell[tl]{arrival\_transport, disposition, hadm\_id\\intime, outtime, stay\_id}\\%
\vspace{0.5em}%
\hspace{1em} pyxis&\makecell[tl]{charttime, name, stay\_id}\\%
\vspace{0.5em}%
\hspace{1em} triage&\makecell[tl]{acuity, dbp, heartrate\\o2sat, pain, resprate\\sbp, stay\_id, temperature}\\%
\vspace{0.5em}%
\hspace{1em} vitalsign&\makecell[tl]{charttime, dbp, heartrate\\o2sat, pain, resprate\\sbp, stay\_id, temperature}\\%
\midrule%
\textbf{hosp}&\\%
\hspace{1em} admissions&\makecell[tl]{admission\_location, admission\_type, admittime\\discharge\_location, dischtime, hadm\_id\\insurance, marital\_status, race}\\%
\vspace{0.5em}%
\hspace{1em} diagnoses\_icd&\makecell[tl]{hadm\_id, icd\_code, icd\_version}\\%
\vspace{0.5em}%
\hspace{1em} drgcodes&\makecell[tl]{description, drg\_code, drg\_type\\hadm\_id}\\%
\vspace{0.5em}%
\hspace{1em} emar&\makecell[tl]{charttime, emar\_id, event\_txt\\hadm\_id, medication}\\%
\vspace{0.5em}%
\hspace{1em} hcpcsevents&\makecell[tl]{chartdate, hadm\_id, short\_description}\\%
\vspace{0.5em}%
\hspace{1em} labevents&\makecell[tl]{charttime, hadm\_id, itemid\\valuenum, valueuom}\\%
\vspace{0.5em}%
\hspace{1em} omr&\makecell[tl]{chartdate, result\_name, result\_value}\\%
\vspace{0.5em}%
\hspace{1em} patients&\makecell[tl]{dod, gender}\\%
\vspace{0.5em}%
\hspace{1em} procedures\_icd&\makecell[tl]{chartdate, hadm\_id, icd\_code\\icd\_version}\\%
\vspace{0.5em}%
\hspace{1em} transfers&\makecell[tl]{careunit, eventtype, hadm\_id\\intime}\\%
\midrule%
\textbf{icu}&\\%
\hspace{1em} icustays&\makecell[tl]{first\_careunit, hadm\_id, intime\\last\_careunit, outtime, stay\_id}\\%
\bottomrule%
\end{tabular}%
\end{table}

\begin{table}
\centering
\caption{Side‑by‑side view of selected columns from the original sample tables (sourced from MIMIC-IV-DEMO) compared with the format of the tokenized timelines in ETHOS.}
\label{tab:original_vs_tokenized}
\begin{tabular}{|ccc|ccc|}
\hline
\multicolumn{3}{|c|}{\textbf{Original data}} & \multicolumn{3}{c|}{\textbf{Tokenized data}}\\ \hline
\multicolumn{3}{|l|}{\textbf{patients.csv}} & \multicolumn{3}{|l|}{\textbf{timeline.csv}} \\
subject\_id & gender & anchor\_age & subject\_id & time & code \\ \hline
10038081 & F & 63 & 10000248 & 2192‑11‑29 18:44:00 & ED\_REGISTRATION \\
10019917 & M & 44 & 10000248 & 2192‑11‑29 18:44:00 & ED\_TRANSPORT//WALK\_IN \\
10019568 & F & 59 & 10000248 & 2192‑11‑29 18:44:00 & ED\_ACUITY \\
10031404 & F & 82 & 10000248 & 2192‑11‑29 18:44:00 & Q2 \\ 
10008287 & F & 43 & 10000248 & 2192‑11‑29 19:03:00 & 15m‑45m \\ \cline{1-3}
\multicolumn{3}{|l|}{\textbf{admissions.csv}} & 10000248 & 2192‑11‑29 19:03:00 & LAB//51146//\% \\
subject\_id & hadm\_id & admittime & 10000248 & 2192‑11‑29 19:03:00 & Q3 \\ \cline{1-3}
10035631 & 22732862 & 2112‑11‑10 15:55:00 & 10000248 & 2192‑11‑29 19:03:00 & LAB//51200//\% \\
10020786 & 23488445 & 2189‑06‑09 12:45:00 & 10000248 & 2192‑11‑29 19:03:00 & Q7 \\
10020187 & 26842957 & 2170‑02‑24 00:00:00 & 10000248 & 2192‑11‑29 19:03:00 & LAB//51221//\% \\
10005866 & 27167814 & 2148‑03‑10 16:16:00 & 10000248 & 2192‑11‑29 19:03:00 & Q4 \\ 
10002428 & 28676446 & 2157‑07‑16 04:09:00 & 10000248 & 2192‑11‑29 19:03:00 & LAB//51222//G/DL \\ \cline{1-3}
\multicolumn{3}{|l|}{\textbf{diagnoses.csv}} & 10000248 & 2192‑11‑29 19:03:00 & Q5 \\ 
subject\_id & hadm\_id & icd\_code & 10000248 & 2192‑11‑29 19:03:00 & LAB//51144//\% \\ \cline{1-3}
10016742 & 29281842 & N179 & 10000248 & 2192‑11‑29 19:04:00 & Q6 \\ 
10023117 & 21607814 & I428 & 10000248 & 2192‑11‑29 19:37:00 & 15m-45m \\
10040025 & 25933959 & I130 & 10000248 & 2192‑11‑29 19:37:00 & HOSPITAL\_ADMISSION \\ 
10014354 & 27494880 & Z955 & 10000248 & 2192‑11‑29 19:37:00 & OBSERVATION \\ 
10014354 & 27487226 & I5033 & 10000248 & 2192‑11‑29 19:37:00 & INSURANCE\_MEDICAID \\ \hline
\end{tabular}
\end{table}

\begin{table}
\centering
\caption{\textbf{Estimated energy consumption of training ETHOS compared to training large language models.} ETHOS is a dedicated model specifically designed for the electronic health records (EHR) domain, which allows it to be substantially smaller and more efficient to train than general-purpose large language models. With only 45 million parameters, ETHOS was trained on 8 A100 GPUs over 46 hours, consuming an estimated 220 kWh of energy. In contrast, universal LLMs such as GPT-3 (175B parameters), LLaMA 3 (8B), and Falcon (40B) require orders of magnitude more compute and energy, consuming between 307,000 and over 1.2 million kWh. 
}
\label{tab:energy-comparison}
\begin{tabular}{lccccc}
\hline
\textbf{Model Name} & \textbf{Params (B)} & \textbf{GPUs Used} & \textbf{Duration} & \textbf{Est. GPU-hours} & \textbf{Est. Energy (kWh)} \\
\hline
ETHOS               & 0.045              & 8×A100             & 46 hours          & 368                      & $\sim$220                 \\
GPT-3               & 175                & 1024×V100          & $\sim$34 days     & $\sim$835,000            & $\sim$1,287,000           \\
LLaMA 3 (8B)        & 8                  & $\sim$512×A100     & $\sim$25 days     & $\sim$307,000            & $\sim$490,000             \\
Falcon (40B)        & 40                 & 384×A100           & $\sim$21 days     & $\sim$193,000            & $\sim$307,000             \\
\hline
\end{tabular}
\end{table}

\begin{figure}
    \centering
    \caption{\textbf{ROC Curves for ETHOS Across All Prediction Tasks.} ROC curves and corresponding area under the curve (AUC) values with 95\% confidence intervals are shown for seven prediction tasks: Hospital Mortality, ICU Admission, Prolonged Stay (>10 days), Composite Outcome (Hospital Mortality + ICU Admission + Prolonged Stay), Hospitalization at Triage, Critical Outcome Within 12h at Triage, and Emergency Department (ED) Re-presentation Within 72h. Each plot includes the fitted ROC curve (orange), unique thresholds (crosses), and the 95\% confidence interval (gray shading). ETHOS demonstrates high predictive performance across all tasks, with AUC values ranging from 0.740 (ED Re-presentation) to 0.936 (Hospital Mortality).}
    \includegraphics[width=\textwidth]{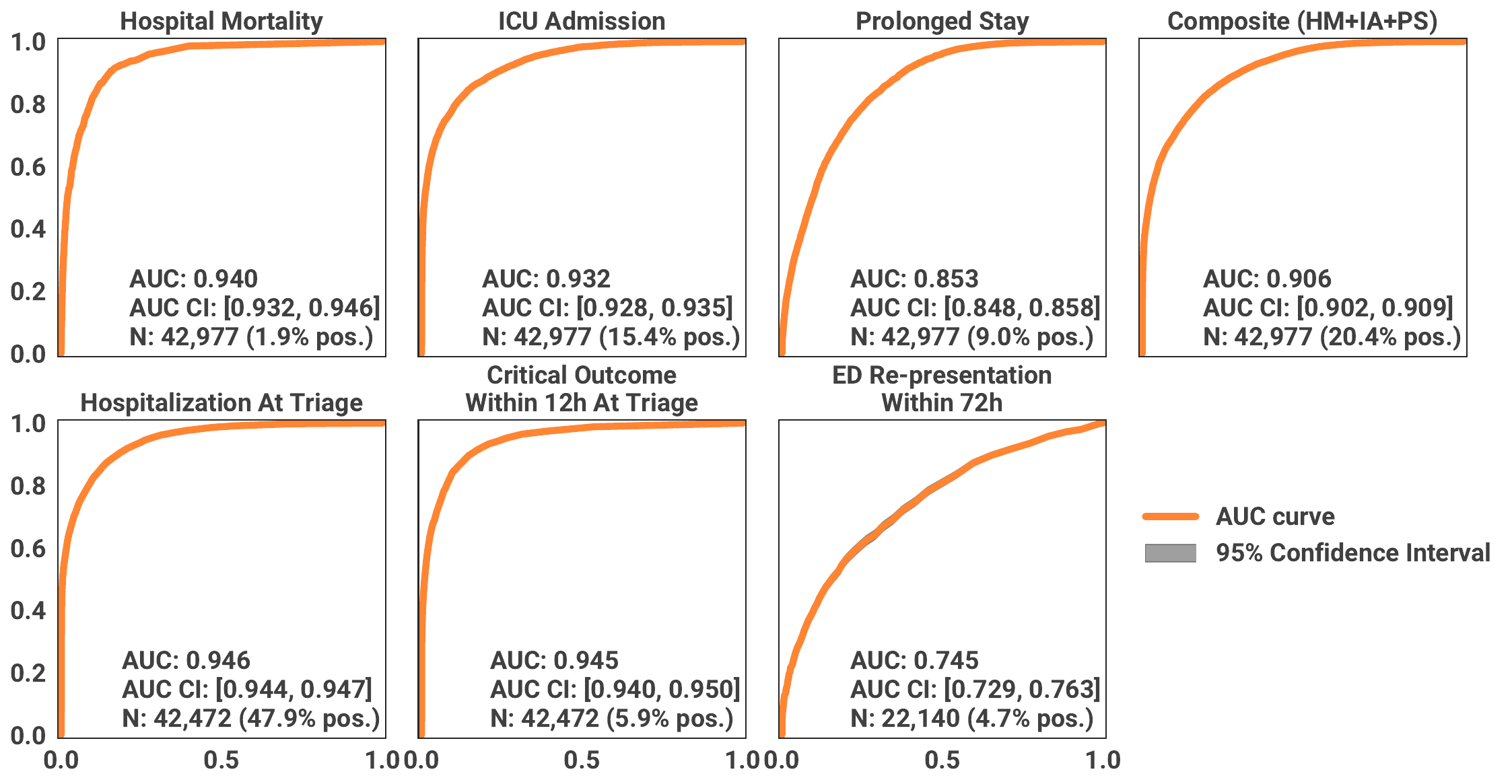}
    \label{fig:ethos-auroc}
\end{figure}

\begin{figure}
    \centering
    \caption{\textbf{AUPRC Curves for ETHOS Across All Prediction Tasks.} Precision-recall (PR) curves and corresponding area under the precision-recall curve (AUPRC) values are shown for seven prediction tasks: Hospital Mortality, ICU Admission, Prolonged Stay (>10 days), Composite Outcome (Hospital Mortality + ICU Admission + Prolonged Stay), Hospitalization at Triage, Critical Outcome Within 12h at Triage, and Emergency Department (ED) Re-presentation Within 72h. Each plot includes the PR curve (orange) and unique thresholds (crosses). ETHOS shows good precision-recall performance across several tasks, with AUPRC values ranging from 0.199 (ED Re-presentation) to 0.887 (Hospitalization at Triage), reflecting the class imbalance present in each task.}

    \includegraphics[width=\textwidth]{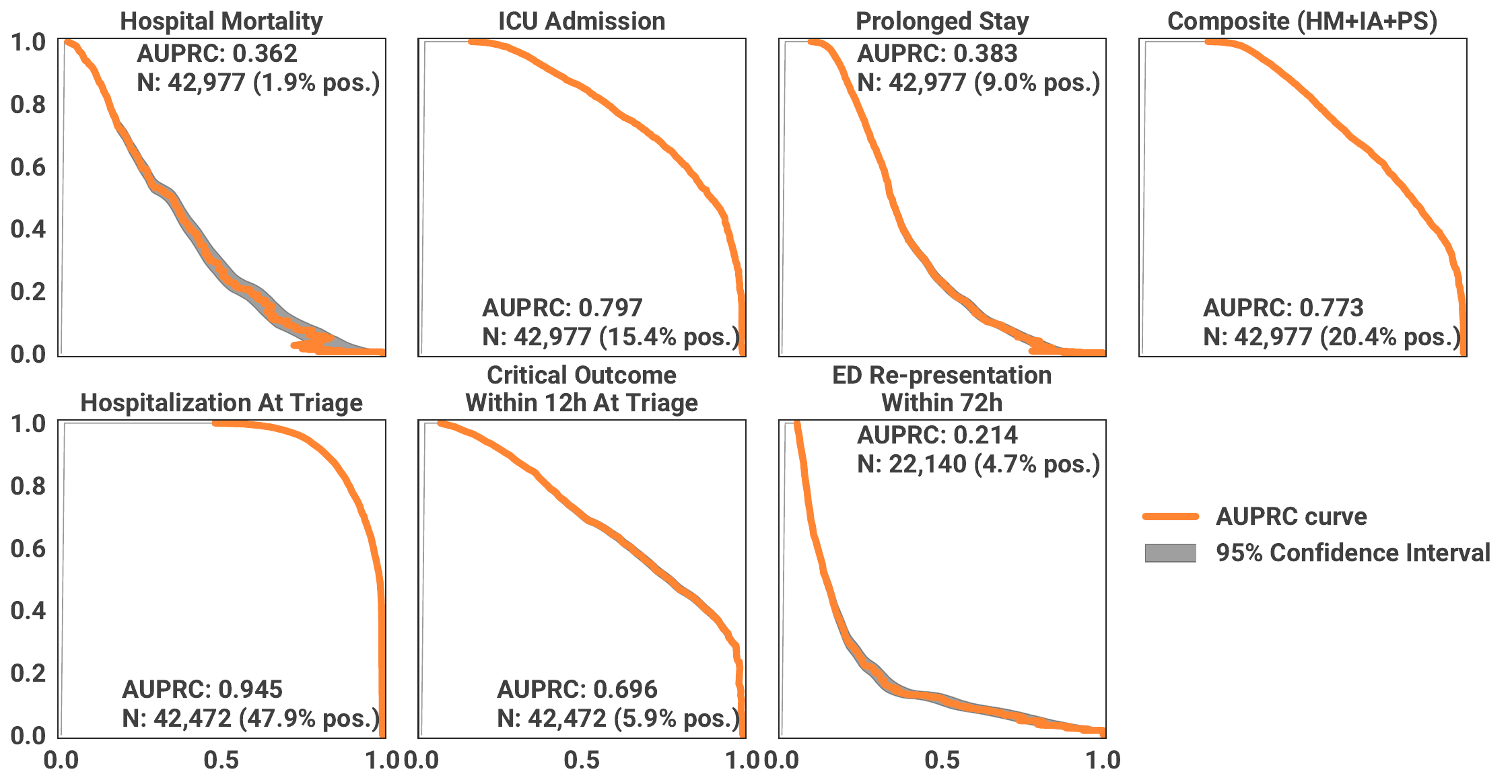}
    \label{fig:ethos-auprc}
    \end{figure}
    
    \begin{figure}
        \centering
        \caption{\textbf{Calibration Curves for ETHOS Predictions Across Clinical Outcomes with 95\% Confidence Intervals Determined by Bootstrapping.} This figure presents calibration curves evaluating the reliability of ETHOS probability predictions across six key clinical outcomes: hospital mortality, ICU admission, prolonged hospital stay, composite risk score (HM+IA+PS), hospitalization at triage, critical outcome within 12 hours at triage, and ED re-presentation within 72 hours. The calibration curves compare predicted probabilities (x-axis) against observed event frequencies (y-axis), with perfect calibration represented by the dashed diagonal line, while the solid orange line shows ETHOS calibration performance, and the shaded gray region represents the 95\% confidence interval (CI) derived from bootstrapping. Each plot includes the Brier score, a metric assessing probabilistic prediction accuracy, where lower values indicate better calibration, with 0.00–0.05 classified as excellent, 0.05–0.10 as good, 0.10–0.20 as acceptable, and values above 0.20 as poor calibration. ETHOS demonstrates excellent calibration for hospital mortality (Brier score: 0.014), critical outcome within 12 hours (0.031), and ED re-presentation (0.041), while ICU admission (0.064), prolonged stay (0.067), and the composite risk score (0.090) exhibit good calibration, closely following the ideal calibration curve. Hospitalization at triage (0.143) is categorized as acceptable calibration, with some deviations at higher predicted probabilities, suggesting areas for potential improvement. Overall, ETHOS exhibits strong calibration across most clinical tasks, particularly in predicting mortality, early critical deterioration, and ED re-presentation, with acceptable performance for hospitalization risk at triage. These findings highlight ETHOS’s reliability in translating probability estimates into clinically meaningful risk stratifications, supporting its potential as a robust AI-driven decision support tool for real-time risk prediction and clinical decision-making.}
        \includegraphics[width=\textwidth]{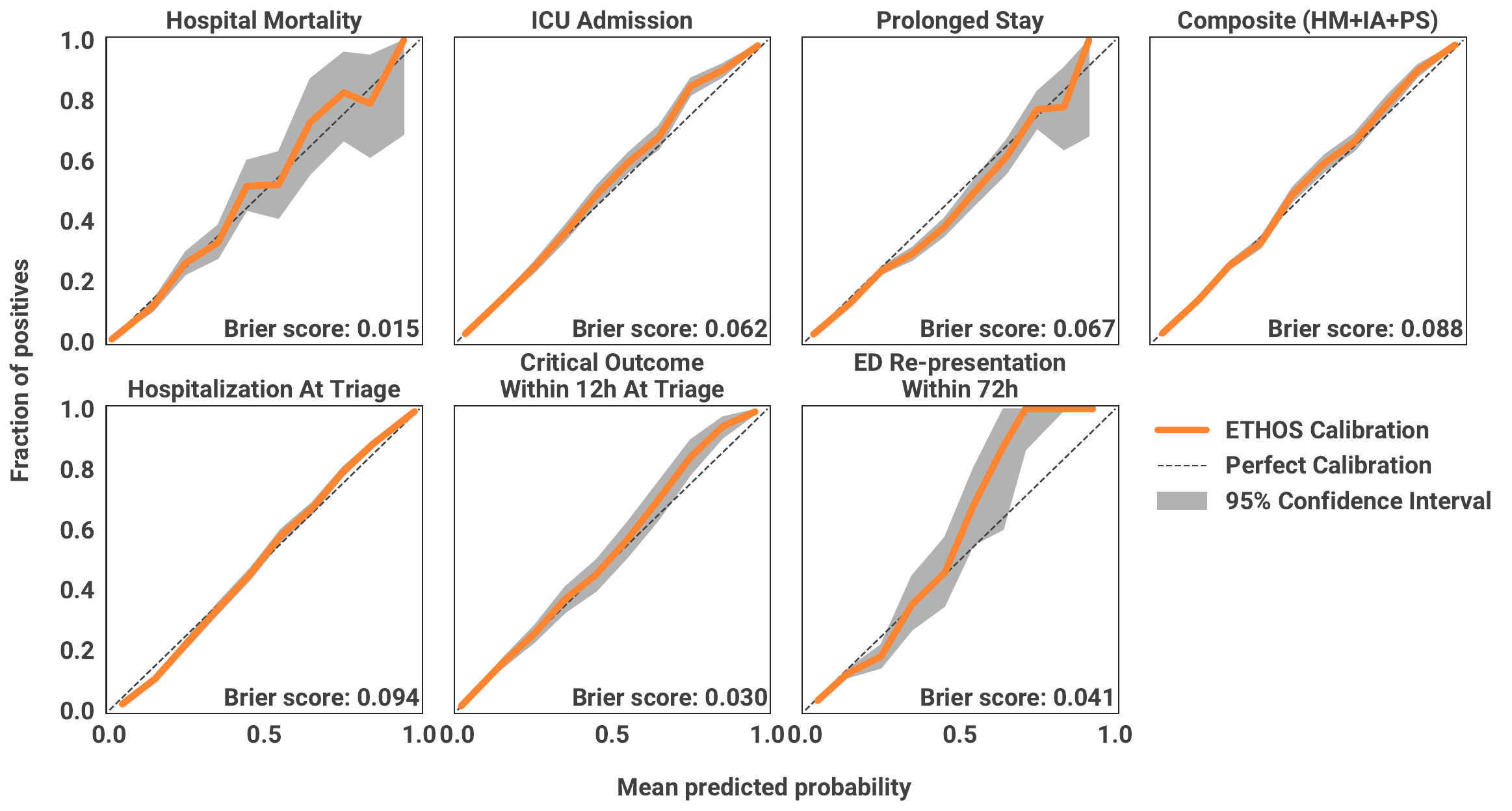}
        \label{fig:ethos-calibration}
\end{figure}

\clearpage
\begin{longtable}{c}
\caption{\textbf{Risk Trajectories for Eight Patients from ED Presentation to Discharge, ICU Admission, or Death.}. This figure presents examples of risk trajectories for eight different patients, illustrating the dynamic evolution of risk predictions following presentation at the emergency department (ED). Each risk value is estimated from multiple (n=100) simulated fPHTs. The shaded area around each risk curve represents the 95\% confidence interval (CI) for the predicted risk. The primary graphs plot risk progression as a function of the number of tokens generated since ED presentation, effectively modeling the temporal evolution of patient risk. The visualisation of ARES score is schematically represented below using 10 color-coded symbols corresponding to key risk categories (see Figures 1 and 2 in main paper). In some graphs, symbols corresponding to ICU admission risk are absent (e.g., E, F, G, and H) because these patients were already admitted to the ICU earlier, leading ARES to automatically exclude this risk component from consideration. The time axis under ARES represents actual elapsed time (in hours and days) since ED presentation. However, time progression on these axes is not linear, as the number of generated tokens does not directly correspond to real-time intervals. Instead, token generation occurs in discrete units determined by patient events. Notably, in case H, a sudden drop in prolonged stay risk occurs because ARES automatically reclassifies a risk of prolonged stay >10 days into prolonged stay >15 days, leading to an observed risk reduction. This drop is an inherent property of ARES modeling rather than a true change in patient status. All trajectories ultimately conclude when the patient either dies or is discharged.} 
\label{tab:timeline-examples} \\

\includegraphics[width=1\textwidth]{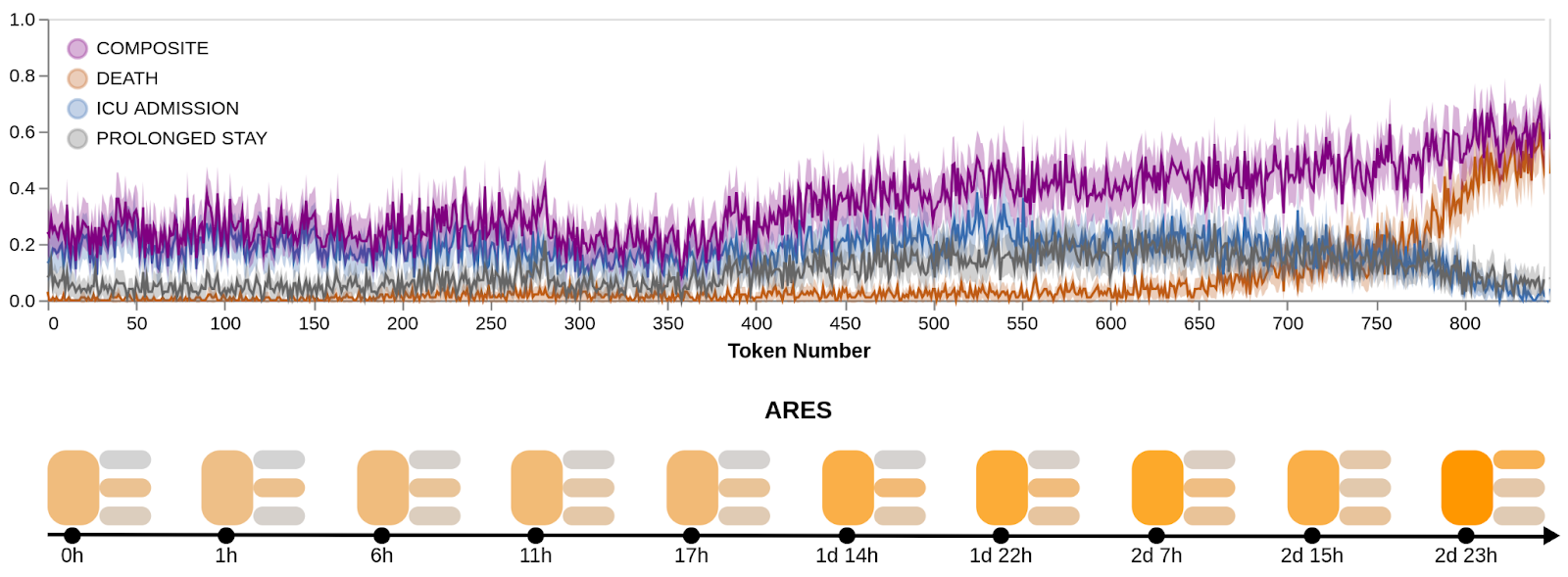} \\
\textbf{(A) Ends with death.} \\[10pt]

\includegraphics[width=1\textwidth]{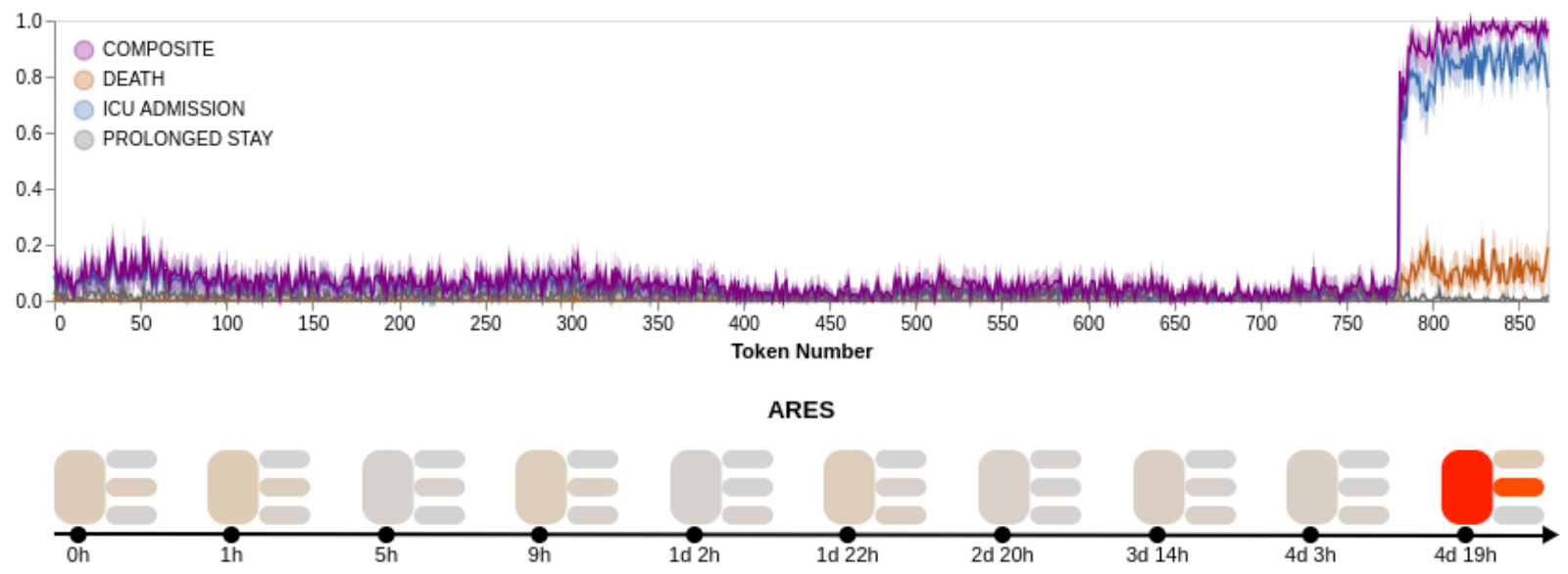} \\
\textbf{(B) Ends with ICU admission.} \\[10pt]

\includegraphics[width=1\textwidth]{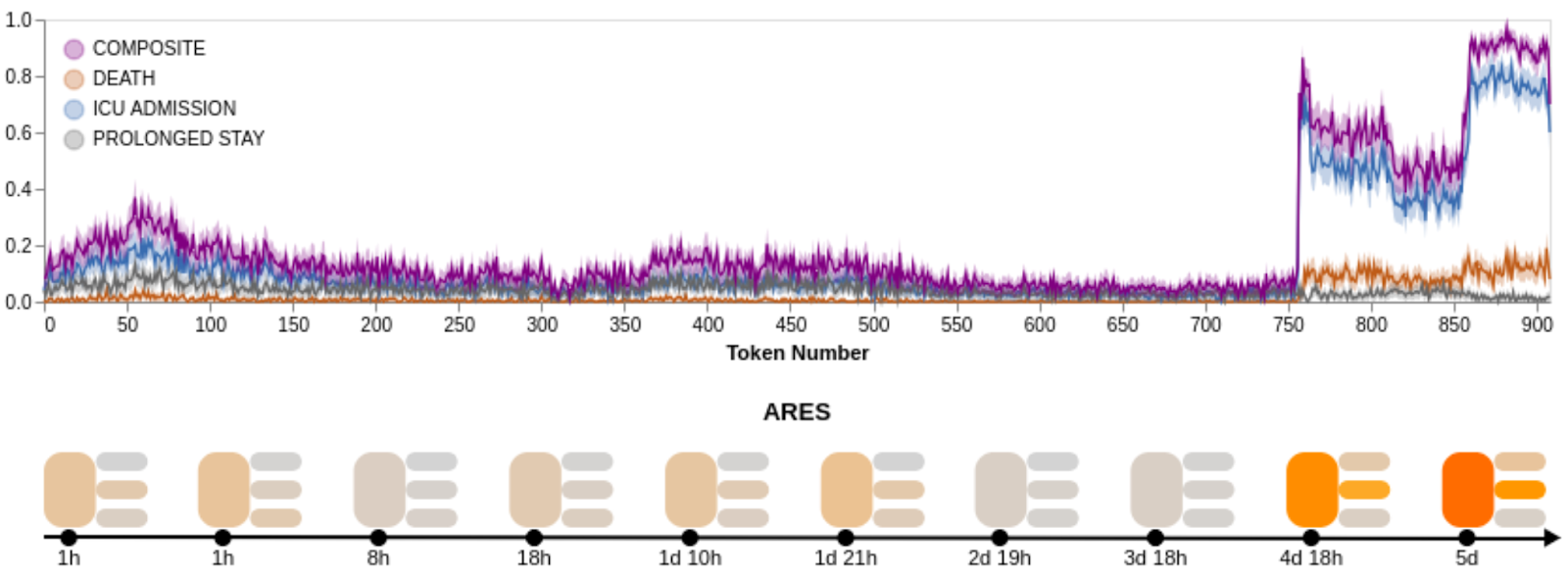} \\
\textbf{(C) Ends with ICU admission.} \\[10pt]

\includegraphics[width=1\textwidth]{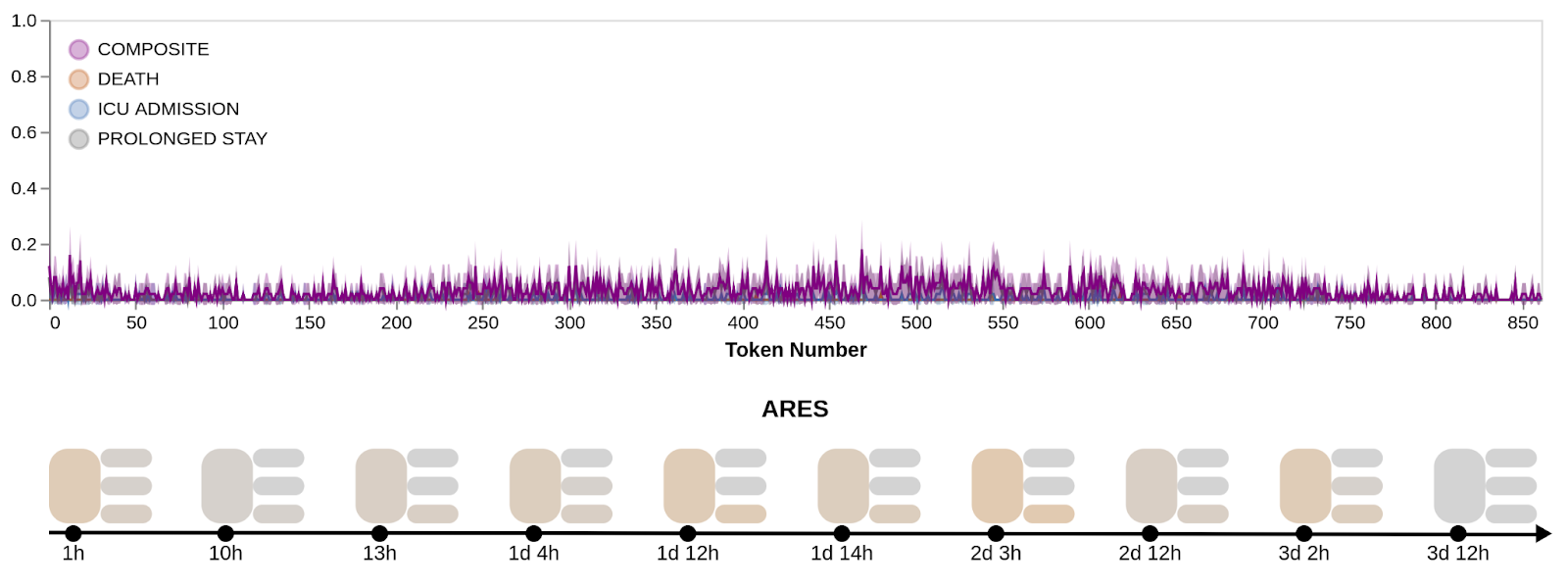} \\
\textbf{(D) Ends with hospital discharge.} \\[10pt]

\includegraphics[width=1\textwidth]{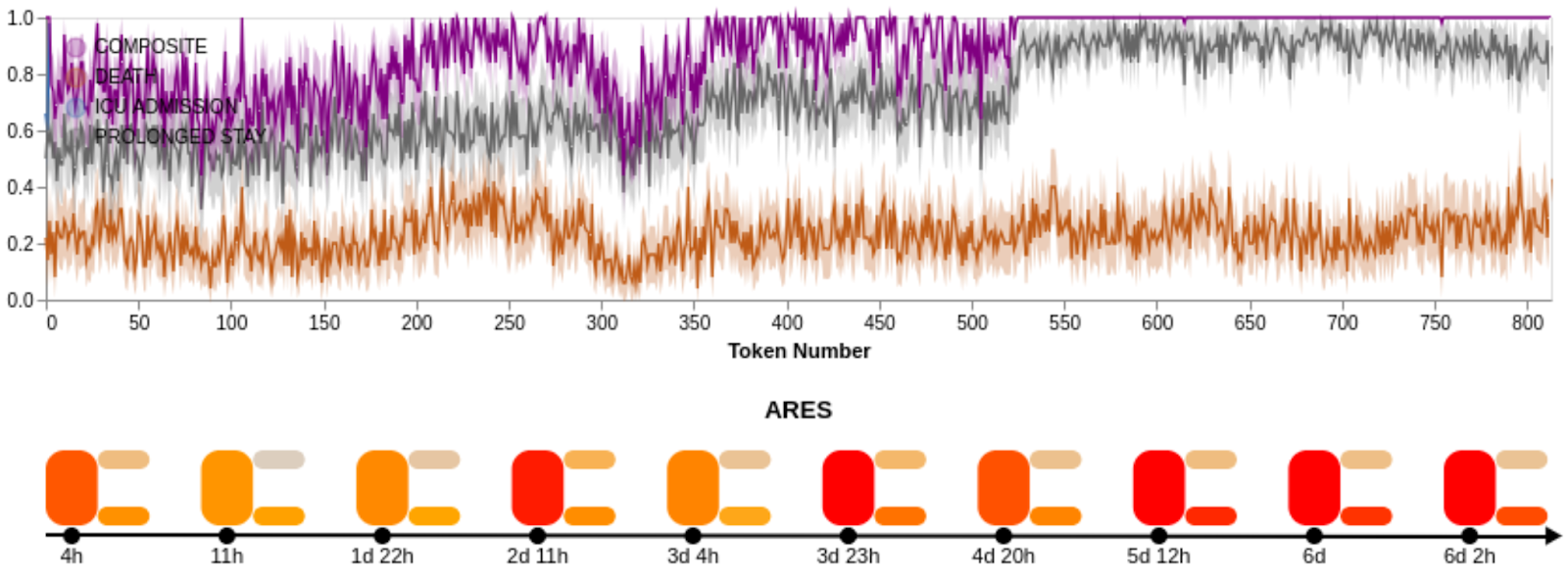} \\
\textbf{(E) Ends with death.} \\[10pt]

\includegraphics[width=1\textwidth]{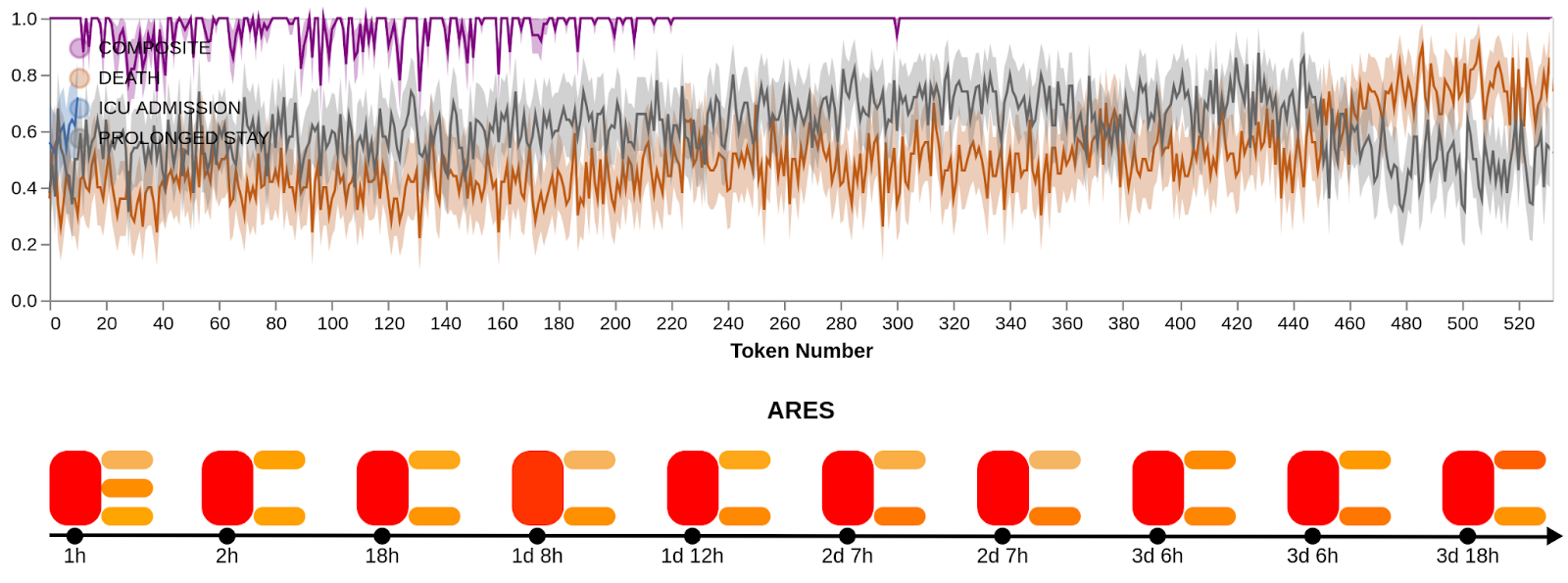} \\
\textbf{(F) Ends with death.} \\[10pt]

\includegraphics[width=1\textwidth]{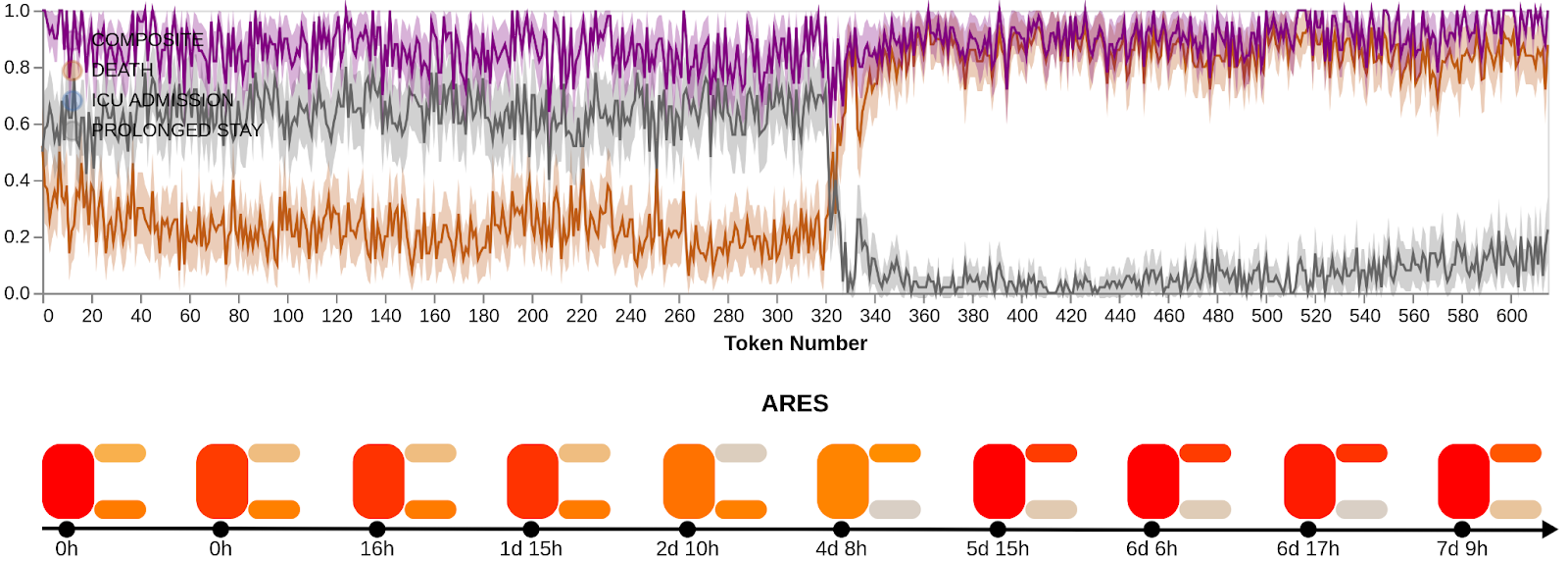} \\
\textbf{(G) Ends with death.} \\[10pt]

\includegraphics[width=1\textwidth]{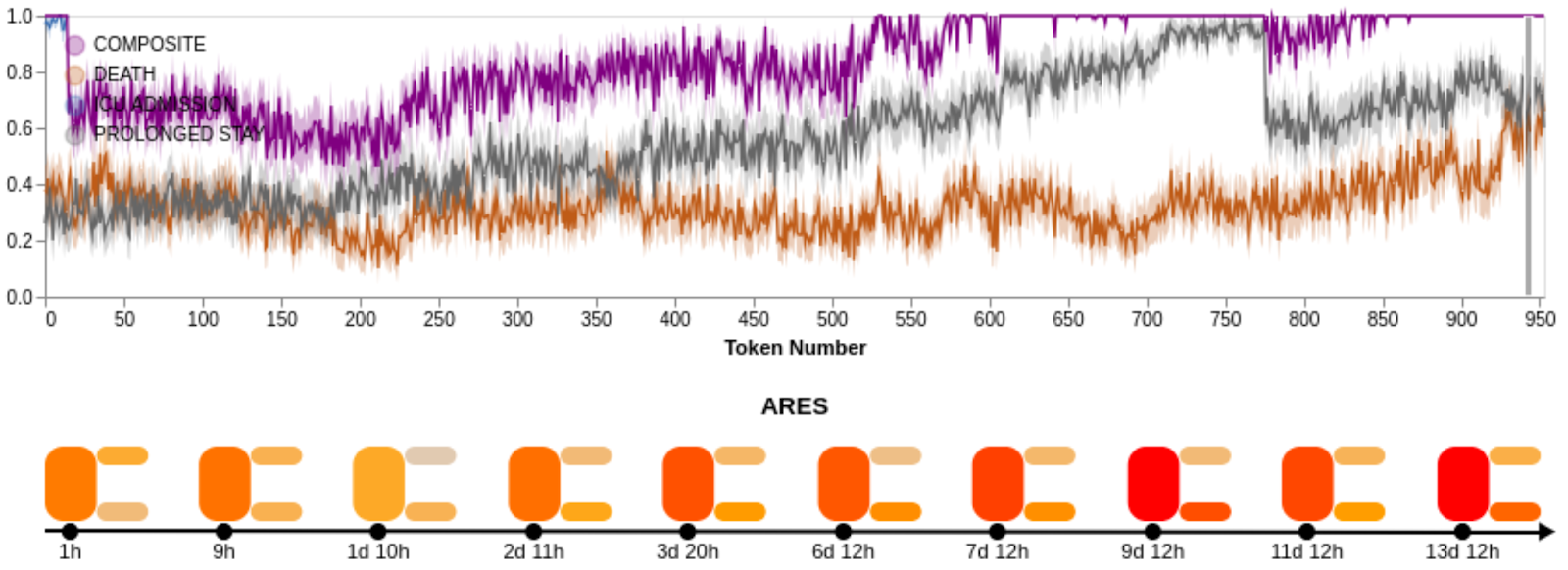} \\
\textbf{(H) Ends with death.} \\[10pt]

\end{longtable}

\begin{longtable}{lcccccc}
\caption{\textbf{Detailed Token Statistics}. The table provides a detailed breakdown of the total number of tokens and unique tokens for each code group in the training, test, and combined datasets. Each code group represents a specific type of information, such as laboratory results (LAB), clinical classifications (e.g., ATC, ICD\_CM), time intervals (e.g., 15m-45m, 12h-18h), and other key features like BMI, vitals, or discharge locations. The statistics summarize the diversity (\#Unique) and frequency (Count) of tokens across datasets, offering insights into the distribution and variability of features used in the modeling process.} \label{tab:detailed-token-stats} \\
\toprule
 & \multicolumn{2}{c}{Train} & \multicolumn{2}{c}{Test} & \multicolumn{2}{c}{Total} \\
 & \#Unique & Count & \#Unique & Count & \#Unique & Count \\
Code Group &  &  &  &  &  &  \\
\midrule
\endfirsthead
\toprule
 & \multicolumn{2}{c}{Train} & \multicolumn{2}{c}{Test} & \multicolumn{2}{c}{Total} \\
 & \#Unique & Count & \#Unique & Count & \#Unique & Count \\
Code Group &  &  &  &  &  &  \\
\midrule
\endhead
\midrule
\multicolumn{7}{r}{Continued on next page} \\
\midrule
\endfoot
\bottomrule
\endlastfoot
LAB & 200 & 90,250,118 & 200 & 10,098,515 & 200 & 100,348,633 \\
ATC & 87 & 26,773,380 & 81 & 2,997,648 & 87 & 29,771,028 \\
ATC\_4 & 12 & 26,773,367 & 12 & 2,997,644 & 12 & 29,771,011 \\
ATC\_SFX & 213 & 26,658,727 & 182 & 2,984,558 & 213 & 29,643,285 \\
Q1 & 1 & 13,313,065 & 1 & 1,476,714 & 1 & 14,789,779 \\
Q2 & 1 & 12,153,214 & 1 & 1,353,936 & 1 & 13,507,150 \\
Q3 & 1 & 11,631,525 & 1 & 1,299,028 & 1 & 12,930,553 \\
Q4 & 1 & 10,483,733 & 1 & 1,172,049 & 1 & 11,655,782 \\
Q5 & 1 & 10,315,908 & 1 & 1,156,166 & 1 & 11,472,074 \\
Q6 & 1 & 10,154,034 & 1 & 1,141,348 & 1 & 11,295,382 \\
VITAL & 6 & 9,946,752 & 6 & 1,113,072 & 6 & 11,059,824 \\
Q7 & 1 & 9,574,210 & 1 & 1,076,334 & 1 & 10,650,544 \\
ICD\_CM & 2,989 & 9,330,094 & 2,542 & 1,036,475 & 2,989 & 10,366,569 \\
Q8 & 1 & 8,954,426 & 1 & 1,006,563 & 1 & 9,960,989 \\
Q9 & 1 & 8,593,863 & 1 & 966,320 & 1 & 9,560,183 \\
Q10 & 1 & 7,900,178 & 1 & 888,383 & 1 & 8,788,561 \\
ICD\_PCS & 34 & 3,998,316 & 34 & 442,617 & 34 & 4,440,933 \\
15m-45m & 1 & 2,234,231 & 1 & 251,165 & 1 & 2,485,396 \\
1h15m-2h & 1 & 2,082,216 & 1 & 232,659 & 1 & 2,314,875 \\
2h-3h & 1 & 1,925,854 & 1 & 214,816 & 1 & 2,140,670 \\
3h-5h & 1 & 1,877,497 & 1 & 209,154 & 1 & 2,086,651 \\
45m-1h15m & 1 & 1,678,348 & 1 & 188,677 & 1 & 1,867,025 \\
5m-15m & 1 & 1,549,919 & 1 & 173,374 & 1 & 1,723,293 \\
BMI & 10 & 1,485,790 & 10 & 169,939 & 10 & 1,655,729 \\
5h-8h & 1 & 1,122,479 & 1 & 124,573 & 1 & 1,247,052 \\
8h-12h & 1 & 980,545 & 1 & 109,975 & 1 & 1,090,520 \\
TRANSFER & 38 & 750,441 & 38 & 83,393 & 38 & 833,834 \\
12h-18h & 1 & 708,241 & 1 & 79,051 & 1 & 787,292 \\
2mt-6mt & 1 & 465,225 & 1 & 52,313 & 1 & 517,538 \\
=6mt & 1 & 456,699 & 1 & 50,085 & 1 & 506,784 \\
30d-2mt & 1 & 430,807 & 1 & 48,696 & 1 & 479,503 \\
12d-20d & 1 & 388,256 & 1 & 44,259 & 1 & 432,515 \\
DRG & 772 & 388,255 & 737 & 42,977 & 772 & 431,232 \\
HOSPITAL\_DISCHARGE & 1 & 388,254 & 1 & 42,977 & 1 & 431,231 \\
DISCHARGE\_LOCATION & 10 & 388,254 & 10 & 42,977 & 10 & 431,231 \\
INSURANCE & 3 & 388,254 & 3 & 42,977 & 3 & 431,231 \\
HOSPITAL\_ADMISSION & 1 & 388,254 & 1 & 42,977 & 1 & 431,231 \\
ADMISSION\_TYPE & 3 & 388,254 & 3 & 42,977 & 3 & 431,231 \\
ED\_REGISTRATION & 1 & 382,614 & 1 & 42,473 & 1 & 425,087 \\
ED\_OUT & 1 & 382,614 & 1 & 42,473 & 1 & 425,087 \\
ED\_ACUITY & 1 & 382,614 & 1 & 42,473 & 1 & 425,087 \\
ED\_TRANSPORT & 4 & 382,614 & 4 & 42,473 & 4 & 425,087 \\
20d-30d & 1 & 340,809 & 1 & 38,149 & 1 & 378,958 \\
4d-7d & 1 & 333,877 & 1 & 38,375 & 1 & 372,252 \\
7d-12d & 1 & 328,988 & 1 & 37,916 & 1 & 366,904 \\
1d-2d & 1 & 307,351 & 1 & 34,627 & 1 & 341,978 \\
TIMELINE\_END & 1 & 257,082 & 1 & 28,540 & 1 & 285,622 \\
2d-4d & 1 & 227,549 & 1 & 25,932 & 1 & 253,481 \\
18h-1d & 1 & 225,224 & 1 & 25,242 & 1 & 250,466 \\
HCPCS & 66 & 127,052 & 37 & 13,731 & 66 & 140,783 \\
ICU\_ADMISSION & 1 & 65,816 & 1 & 7,365 & 1 & 73,181 \\
ICU\_TYPE & 9 & 65,816 & 9 & 7,365 & 9 & 73,181 \\
ICU\_DISCHARGE & 1 & 65,816 & 1 & 7,365 & 1 & 73,181 \\
SOFA & 1 & 65,816 & 1 & 7,365 & 1 & 73,181 \\
MEDS\_DEATH & 1 & 26,200 & 1 & 2,876 & 1 & 29,076 \\
\end{longtable}

\section{Monte Carlo Justification for Probability Estimation}
\label{sec:monte-carlo-just}

Let $p(\mathbf{x})$ denote the probability distribution over fPHTs as modeled by ETHOS where by $\mathbf{x}$ we indicate an fPHT. Suppose we want to estimate the probability of some event $A$ regarding the future timeline. For instance, $A$ could be the event ``the patient death when admitted'' or ``the patient admitted to ICU.'' Formally,

\[
\Pr(A) \;=\; \sum_{\mathbf{x} \in A} p(\mathbf{x}),
\]

where the sum is over all sequences $\mathbf{x}$ for which the event $A$ holds (i.e., $\mathbf{x} \in A$).

{\bf A. Monte Carlo Estimator}\\

A straightforward Monte Carlo approach to approximate $\Pr(A)$ is as follows:

\begin{enumerate}
    \item \textbf{Draw} $N$ i.i.d.\ samples $\{\mathbf{x}^{(1)}, \mathbf{x}^{(2)}, \ldots, \mathbf{x}^{(N)}\}$ from the model $p(\mathbf{x})$.
    \item \textbf{Define} an indicator function $I(\mathbf{x}^{(i)} \in A)$, which is $1$ if the sample $\mathbf{x}^{(i)}$ lies in $A$, and $0$ otherwise.
    \item \textbf{Estimate} $\Pr(A)$ by the ratio
    \[
       \hat{\Pr}(A) 
       \;=\; \frac{1}{N} \sum_{i=1}^{N} I\bigl(\mathbf{x}^{(i)} \in A\bigr).
    \]
\end{enumerate}

In other words, $\hat{\Pr}(A)$ is simply the fraction of samples whose corresponding timelines satisfy event $A$ indicated as $M/N$ in the text.

{\bf B. Unbiasedness}\\

If the samples $\mathbf{x}^{(i)}$ are drawn exactly from $p(\mathbf{x})$, then for each sample,

\[
\mathbb{E}[I(\mathbf{x}^{(i)} \in A)]
\;=\; \Pr(\mathbf{x}^{(i)} \in A)
\;=\; \Pr(A).
\]

Hence,
\[
\mathbb{E}\bigl[\hat{\Pr}(A)\bigr]
\;=\; \mathbb{E}\Bigl[\frac{1}{N}\sum_{i=1}^N I(\mathbf{x}^{(i)} \in A)\Bigr]
\;=\; \Pr(A),
\]
showing that $\hat{\Pr}(A)$ is an \emph{unbiased} estimator of $\Pr(A)$.

{\bf{C. Convergence by the Law of Large Numbers}}\\

By the Law of Large Numbers (LLN), as $N \to \infty$,
\[
\hat{\Pr}(A) \;\xrightarrow{a.s.}\; \Pr(A),
\]
meaning the simple ratio of ``successes'' (i.e., samples satisfying $A$) to total draws converges almost surely to the true probability.

\section{Intuitive Operation of ETHOS}
\label{sec:intuitive-ethos}

\autoref{fig:model-info} shows the decoder‐only transformer backbone of ETHOS, which follows the standard GPT design.  Below we provide an intuitive, step‐by‐step explanation of how ETHOS works “zero‐shot,” i.e.\ with no additional fine‐tuning for specific prediction tasks.

\paragraph{1. Treating Patient History as a “Story.”}  
Each Patient Health Timeline (PHT) is a long sequence of discrete tokens, think of each token as a “word” in a clinical narrative.  Static tokens (e.g., demographics) set the scene, categorical tokens (e.g., diagnoses) are like domain‐specific vocabulary, and time‐interval tokens keep track of elapsed time between events.  Just as a language model reads a sentence and predicts the next word, ETHOS reads a patient’s history and predicts the next medical event token.

\paragraph{2. Zero‐Shot Generative Inference.}  
Because ETHOS has been trained to model the joint distribution of these tokens, at inference time it can \emph{continue} any PHT without further training.  We provide the model with the tokens of a patient’s past, and then let it generate new tokens one at a time.  No task‐specific labels or fine‐tuning are needed—hence “zero‐shot.”  Each new token corresponds to a plausible future event (e.g., a lab draw, medication change, or hospital admission).

\paragraph{3. Exploring Possible Futures via Sampling.}  
To capture uncertainty, we perform many independent “completions” of the PHT.  Each completion (an fPHT) is analogous to asking, “What could happen next?” and letting the model write the next chapter of the patient’s story.  We use nucleus sampling (top-\(p\)) to allow diversity: at each step, ETHOS randomly selects the next token from the smallest set whose cumulative probability exceeds \(p=0.9\).  By generating \(N\) such fPHTs, we build a Monte Carlo ensemble of future trajectories.

\paragraph{4. From Generated Tokens to Risk Estimates.}  
Some tokens are can be treated as outcome (e.g., mortality).  To compute inpatient mortality, we simply count how many of the \(N\) simulated fPHTs include that token while patient is in hospital (before discharge token).  The fraction \(M/N\) then provides a direct, probabilistic estimate of risk.  This simulation‐based approach naturally accounts for multiple pathways and branching possibilities in a patient’s course.

\paragraph{5. Toward Controlled Cohort Generation.}  
Because ETHOS is a pure generator, one can modify the sampling distribution, reweighting certain token types (e.g., age or comorbidity tokens), to synthesize patient cohorts with desired characteristics.  This capability opens the door to fairness‐aware risk modeling and targeted “what‐if” analyses, which we plan to explore in future work.

\section{Baseline Models}
\label{sec:baselines}
To demonstrate the applicability of ETHOS as the base for ARES we compared its performance on three benchmark tasks against other baseline models. Those tasks were: prediction of the hospital admission at triage, prediction of the critical outcome (death or transfer to ICU within 12 hours) at triage, and ED re-presentation within 72 hours after discharge from ED. We followed the baseline models utilized for the benchmark tasks as presented in~\cite{Xie2022-ur}. Firstly, we used clinically applied scoring systems: Modified Early Warning Score (MEWS)~\cite{Subbe2001-jx}, National Early Warning Score (NEWS, versions 1 and 2)~\cite{Williams2022-ah,Smith2013-jz,Zhang2025-gp}, Rapid Emergency Medicine Score (REMS)~\cite{Olsson2004-zw}, and Cardiac Arrest Risk Triage (CART)~\cite{Churpek2012-ie}. Those scores require the collection of specific clinical features like Heart Rate, Respiratory Rate or Oxygen Saturation. They cannot adapt to the dynamic state of the patient unless all the features used by the scoring system are remeasured. Similarly, Emergency Severity Index (ESI)~\cite{Eitel2003-nq}, a five-level triage system, was assigned by a nurse. In addition, we created custom scores with the AutoScore method that generates scoring systems from clinical features automatically in six steps, including variable ranking and transformation as well as score derivation by weighting and normalization ~\cite{xie2020autoscore}. Next, we utilized classic machine learning algorithms, including Logistic Regression (LR), which is a linear classifier, Multi-layer Perceptron (MLP), a non-linear classifier and tree-based, ensemble algorithms Random Forest (RF) and Gradient Boosting (GB). Finally, we applied more advanced deep learning-based algorithms  Med2Vec~\cite{Choi2016-fe}, which applies non-linear transformations to vector embeddings of ICD codes and Long Short-Term Memory (LSTM)~\cite{Hochreiter1997-af} a neural network accounting for temporal changes in data for the ED re-presentation task. We make our adapted baseline model training code publicly available for reproducibility purposes \footnote{https://github.com/ipolharvard/mimic4ed-benchmark}. 

\section*{Details of model training}

ETHOS was implemented as a decoder-only transformer. Training data were constructed by appending an “End of timeline” token to each patient’s PHT and concatenating all timelines into one long token sequence. The model was then trained, in an unsupervised fashion, to predict the next token given its preceding context by minimizing the standard cross-entropy loss, exactly as in large-scale language-model pretraining. Given the scale of our PHT corpus (hundreds of millions of tokens) and the model’s complexity, we matched the parameter count of GPT-2 as a starting point, conducted a focused hyperparameter search (see~\autoref{fig:model-info}), and made heuristic adjustments to optimize convergence. Training was performed across eight NVIDIA GPUs over approximately 36 hours, reflecting resource requirements similar to those for open-domain transformer models. AdamW optimizer (decoupled weight decay) with hyperparameters matched to those in nanoGPT/GPT-2–scale pretraining. In our experiments, we used a constant learning rate of $4e^{-4}$, weight-decay of 0.01, and default momentum parameters ($\beta_1=0.9$ and $\beta_2=0.999$ training for 100K steps per federated round with batch size 32. For more details, refer to~\cite{Renc2024-jf}. All code including training scripts is publicly available in our GitHub repository at  \url{https://github.com/ipolharvard/ethos-ares}.

\end{document}